%% file: main.tex
\documentclass{article} %
\usepackage[table]{xcolor}
\usepackage{iclr2026_conference,times}

\input{math_commands.tex}

\usepackage{hyperref}
\usepackage{url}
\usepackage{xspace}

\usepackage{siunitx} %
\definecolor{turquoise}{cmyk}{0.65,0,0.1,0.3}
\definecolor{purple}{rgb}{0.65,0,0.65}
\definecolor{dark_green}{rgb}{0, 0.5, 0}
\definecolor{orange}{rgb}{0.8, 0.6, 0.2}
\definecolor{red}{rgb}{0.8, 0.2, 0.2}
\definecolor{darkred}{rgb}{0.6, 0.1, 0.05}
\definecolor{blueish}{rgb}{0.0, 0.3, .6}
\definecolor{light_gray}{rgb}{0.7, 0.7, .7}
\definecolor{pink}{rgb}{1, 0, 1}
\definecolor{greyblue}{rgb}{0.25, 0.25, 1}

\definecolor{bestcol}{RGB}{254,196,79}
\definecolor{secondbestcol}{RGB}{255,247,188}

\definecolor{bestoneviewcol}{RGB}{147,129,255}
\definecolor{secondbestoneviewcol}{RGB}{229,229,255}

\newcommand{\bestoneview}[1]{\cellcolor{bestoneviewcol} \textbf{#1}}
\newcommand{\best}[1]{\cellcolor{bestcol} \textbf{#1}}
\newcommand{\secondbest}[1]{\cellcolor{secondbestcol} #1}
\newcommand{\secondbestoneview}[1]{\cellcolor{secondbestoneviewcol} #1}

\usepackage{xspace} %
\newcommand{\titlecaption}[2]{\caption{\textbf{#1.}\xspace#2}}

\definecolor{slowtimecol}{RGB}{230,230,230}

\newcommand{\projectpage}{\href{https://reli3d.jdihlmann.com/}{Project Page: https://reli3d.jdihlmann.com/}}

\newcommand{\shorttitle}{ReLi3D: Relightable Multi-view 3D \mbox{Reconstruction} with Disentangled \mbox{Illumination}}
\newcommand{\ours}{ReLi3D\xspace}

\usepackage{tikz}
\usepackage{adjustbox}
\usepackage{cleveref}
\usepackage{booktabs}

\title{\shorttitle}

\author{
Jan-Niklas Dihlmann\thanks{Also at Stability AI.} \\
University of T\"ubingen \\
\And
Mark Boss \\
Stability AI \\
\And
Simon Donn\'e \\
Stability AI \\
\And
Andreas Engelhardt \\
Stability AI \\
\And
Hendrik P.A.\ Lensch \\
University of T\"ubingen \\
\And
Varun Jampani \\
Stability AI
}

\iclrfinalcopy
\begin{document}

\maketitle

\input{content/00_abstract.tex}

\input{figures/tex/teaser.tex}

\input{content/01_introduction.tex}

\input{content/02_related_work.tex}

\input{content/03_background.tex}

\input{content/04_method.tex}

\input{content/05_experiments.tex}

\input{content/06_conclusion.tex}

\section*{Acknowledgements}
The authors thank Stability AI for hosting Jan-Niklas Dihlmann as an intern during this work. This work was funded by the Deutsche Forschungsgemeinschaft (DFG, German Research Foundation) under Germany's Excellence Strategy---EXC number 2064/1---Project number 390727645. This work was supported by the German Research Foundation (DFG): SFB 1233, Robust Vision: Inference Principles and Neural Mechanisms, TP~02, project number: 276693517. This work was supported by the T\"ubingen AI Center. The authors thank the International Max Planck Research School for Intelligent Systems (IMPRS-IS) for supporting Jan-Niklas Dihlmann.

\bibliography{references,spar3d,spv,lirm}
\bibliographystyle{iclr2026_conference}

\appendix
\input{content/xx_supplements.tex}

\end{document}

%% file: math_commands.tex
\usepackage{amsmath,amsfonts,bm}

\def\eqref#1{equation~\ref{#1}}

\def\1{\bm{1}}

\DeclareMathAlphabet{\mathsfit}{\encodingdefault}{\sfdefault}{m}{sl}
\SetMathAlphabet{\mathsfit}{bold}{\encodingdefault}{\sfdefault}{bx}{n}

%% file: content/00_abstract.tex
\begin{abstract}
    
    Reconstructing 3D assets from images has long required separate pipelines for geometry reconstruction, material estimation, and illumination recovery, each with distinct limitations and computational overhead. 
    We present \ours, the first unified end-to-end pipeline that simultaneously reconstructs complete 3D geometry, spatially-varying physically-based materials, and environment illumination from sparse multi-view images in under one second.
    Our key insight is that multi-view constraints can dramatically improve material and illumination disentanglement, a problem that remains fundamentally ill-posed for single-image methods. 
    Key to our approach is the fusion of the multi-view input via a transformer cross-conditioning architecture, followed by a novel unified two-path prediction strategy. 
    The first path predicts the object's structure and appearance, while the second path predicts the environment illumination from image background or object reflections.
    This, combined with a differentiable Monte Carlo multiple importance sampling renderer, creates an optimal illumination disentanglement training pipeline.
    In addition, with our mixed domain training protocol, which combines synthetic PBR datasets with real-world RGB captures, we establish generalizable results in geometry, material accuracy, and illumination quality.
    By unifying previously separate reconstruction tasks into a single feed-forward pass, we enable near-instantaneous generation of complete, relightable 3D assets. \projectpage{}
\end{abstract}

%% file: figures/tex/teaser.tex
\begin{figure}[ht]
    \centering
    \includegraphics[width=0.99\textwidth]{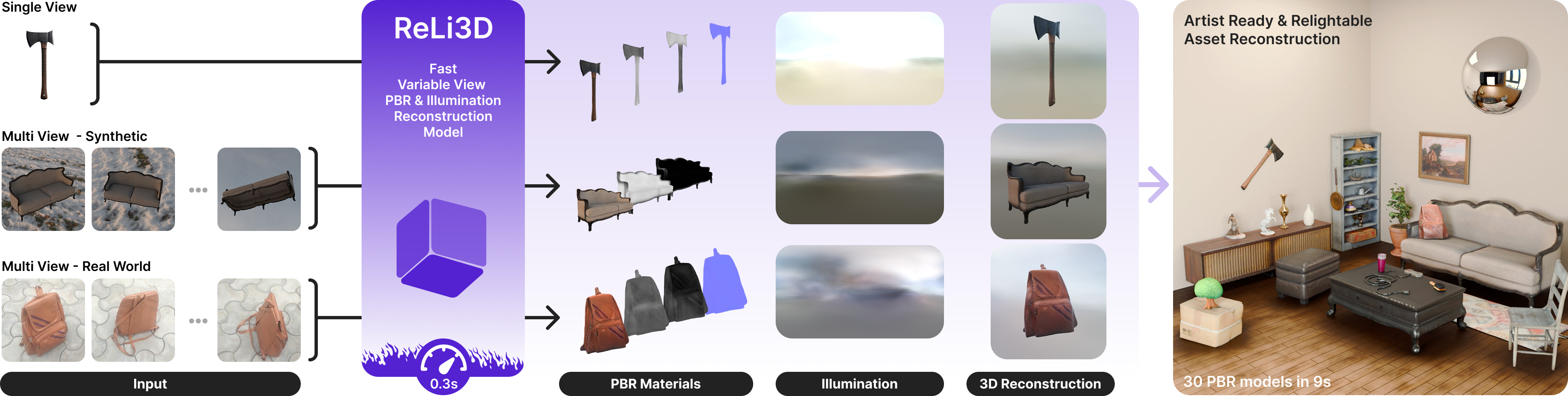}
    \titlecaption{Fast, illumination disentangled reconstructions}{\ours reconstructs high-quality 3D meshes with physically based materials from sparse input images, while disentangling illumination effects; all in just 0.3s.
    It is robustly trained on cross-domain datasets and excels in both single- and multi-view cases, on synthetic data as well as on real-world examples.
    }
    \label{fig:teaser}
\end{figure}

%% file: content/01_introduction.tex
\begin{section}{Introduction}
    \label{sec:introduction}
    
    Reconstructing production-ready 3D assets from images remains a challenging task with immense potential for industrial design, interactive media, or robotics. Two lines of progress have emerged: (i) Generative models based on diffusion, which can achieve striking geometric fidelity, but with long inference times and hallucination, (ii) Large Reconstruction Models (LRMs) such as LRM~\citep{hong2023lrm}, SF3D~\citep{sf3d2024}, and TripoSR~\citep{TripoSR2024} that perform direct feed-forward inference from images to 3D.
    While LRMs are fast and practical, a gap persists between research prototypes and what artists require from a 3D reconstruction, which is accurate reconstruction from multiple views and illumination disentanglement resulting in spatially varying Physically Based Rendering (PBR) materials that support relighting.

    Unfortunately, many existing approaches optimize only for single-view reconstruction, which is inherently ill-posed.
    The same 2D appearance can arise from numerous combinations of surface reflectance and illumination. 
    Regularization or learned priors help, but ambiguity remains, especially in unobserved areas, leading to incomplete spatially varying material predictions, unreliable normals, and therefore limited relighting fidelity. 

    In our perspective, geometric consistency across multiple views provides the missing constraints to separate material properties from lighting effects. 
    When multiple observations see the same surface point under a common illumination, cross-view agreement narrows the feasible solution space and turns an ill-posed single-view problem into a much better constrained one. 
    To operationalize this, we design an architecture where multi-view fusion is not an add-on for robustness, but the primary mechanism for material-lighting disentanglement.

    In this paper, we present \ours, a unified feed-forward system that turns a variable number of posed images into a textured mesh with spatially varying PBR materials and a coherent HDR environment in less than a second. 
   In order to allow for \textbf{Multiview Illumination Disentanglement Reconstruction} we utilize a two-path approach achieved through the following novel contributions:
    \begin{itemize}
        \item \textbf{Cross-view Fusion} A shared cross-conditioning transformer ingests an arbitrary number of views and builds unified feature triplanes used by both paths, driving consistency across viewpoints.
        \item \textbf{Two-path Illumination Disentanglement.} A \emph{geometry+appearance path} yields mesh and svBRDF (albedo/roughness/metallic/normal) from this unified triplane, while a \emph{lighting path} fuses mask-aware tokens to predict an efficient RENI++~\citep{gardner2023reni++} latent code representing a coherent HDR environment. 
        \item \textbf{Disentangled Training via MC+MIS.} A differentiable physically-based Multiple Importance Sampling (MIS) Monte Carlo (MC) renderer ties both paths together, enforcing physically meaningful materials and illumination disentanglement.
        \item \textbf{Mixed-domain Training.} We train on a mixture of synthetic PBR-supervised data and real multi-view captures using image space self-supervision to bridge the gap and allow for real-world generalization.
    \end{itemize}
    
    Together, these pieces deliver the first feed-forward pipeline that jointly reconstructs geometry, spatially varying materials, and HDR illumination at interactive speed. 
    Our experiments show improved reconstruction, relighting fidelity and material realism over recent (i) generative and (ii) reconstruction pipelines; we will release code and weights to foster adoption and reproducibility.

\end{section}

%% file: content/02_related_work.tex
\begin{section}{Related Work}
    \label{sec:related_work}
    \ours lies at the intersection of 3D reconstruction, inverse rendering, and appearance estimation.
    The most closely aligned approaches are image-to-3D reconstruction and generation methods, and we seek to clearly differentiate our feed-forward approach from optimization-based reconstruction methods.

    \begin{paragraph}{Inverse Rendering}
        \label{sec:related_work:inverse_rendering}
        Inverse rendering estimates shape, appearance, and environment lighting from image observations, an inherently ambiguous problem with many plausible material-lighting combinations explaining identical observations.
        Modern methods leverage differentiable rendering~\citep{li2018differentiable,liu2019soft} with scene representations such as NeRF~\citep{mildenhall2021nerf} or Gaussian splats~\citep{kerbl20233d} to reconstruct scenes from dense RGB imagery~\citep{zhang2021nerfactor,boss2021nerd,boss2022samurai,engelhardt2024shinobi,liang2024gs,dihmann2024sssgs}.
        Although regularization losses in shape, materials, or environment~\citep{barron2013intrinsic,li2018materials,gardner2017learning} help reduce ambiguity, these optimization-based approaches require dense multi-view imagery and lengthy inference times.
        None manages to reconstruct 3D objects from sparse views, let alone single images.
        In contrast, \ours performs feed-forward inference from sparse views while jointly estimating spatially varying materials and HDR environments via RENI++~\citep{gardner2023reni++}.
    \end{paragraph}

    \begin{paragraph}{Image-to-3D Generation}
        \label{sec:related_work:generation}
        Score Distillation Sampling methods~\citep{poole2022dreamfusion,shi2023mvdream,wang2024prolificdreamer} optimize 3D representations using 2D diffusion priors but suffer from artifacts and impractically slow inference.
        Multi-view generation approaches~\citep{liu2023zero,long2023wonder3d,voleti2024sv3d,tang2024lgm} first generate consistent views and then apply reconstruction, but face view inconsistencies and inherit inverse rendering ambiguities.
        
        Direct 3D diffusion methods model object distributions in triplane~\citep{shue20233d,cheng2023sdfusion,yariv2024mosaic} or compressed latent spaces~\citep{zhao2025hunyuan3d,xiang2024trellis}.
        SPAR3D~\citep{huang2025spar3d} uniquely diffuses both geometry and PBR materials by first generating sparse point clouds and then regressing detailed structure and appearance, but requires expensive probabilistic sampling.
        The lack of large-scale PBR data typically precludes joint geometry-material modeling in diffusion frameworks.
        Our feed-forward approach achieves comparable quality without the computational overhead of generative sampling, enabling end-to-end joint structure and appearance prediction.
    \end{paragraph}

    \begin{paragraph}{Image-to-3D Reconstruction}
        \label{sec:related_work:reconstruction}
        Early regression approaches~\citep{choy20163d,wang2018pixel2mesh,mescheder2019occupancy} were limited by small datasets like ShapeNet~\citep{chang2015shapenet}, restricting generalization.
        Large Reconstruction Models (LRMs)~\citep{hong2023lrm,tochilkin2024triposr,sf3d2024} now perform direct feed-forward inference at scale using transformer architectures and large datasets~\citep{deitke2022objaverse,reizenstein2021co3d}.
        
        Although fast and practical, existing methods such as SF3D~\citep{sf3d2024} predict only single roughness/metallic values per object rather than spatially varying materials, and lack environment estimation.
        Most critically, these approaches optimize for single-view reconstruction, leaving material-lighting disentanglement fundamentally ill-posed, and the same appearance can arise from countless material-illumination combinations.
        
        The parallel work LIRM~\citep{li2025lirm} addresses similar goals through progressive optimization but lacks illumination prediction and relies purely on synthetic supervision, limiting real-world applicability.
        \ours uniquely leverages multi-view constraints as the primary mechanism for material-lighting disentanglement, enabling robust spatially varying PBR reconstruction with environment estimation through mixed-domain training that bridges synthetic and real-world data.
    \end{paragraph}
\end{section}

%% file: content/03_background.tex
\begin{section}{Preliminaries}
    \label{sec:background}
    
    Reconstructing 3D objects with realistic materials and lighting from images requires understanding how light interacts with surfaces and how to efficiently represent 3D information. This section introduces the key concepts underlying our approach: physically based material representations, environment illumination modeling, and neural 3D representations that enable feed-forward reconstruction.

    \begin{subsection}{Physically Based Material Representation}
        \label{sec:background:pbr}
        An object's visual appearance results from how its surface reflects and refracts light, formally described by the bidirectional reflectance distribution function (BRDF) $f_r(\omega_\text{in}, \omega_\text{out})$.
        This function models the fraction of light reflected into direction $\omega_\text{out}$ given incoming light from direction $\omega_\text{in}$.
        When material properties vary across the surface, we have a spatially varying BRDF (svBRDF).
        
        In practice, we parameterize materials using Disney's principled BRDF~\citep{burley2012disneybrdf} with metallic-roughness representation: RGB albedo (base color) $\rho$, scalar roughness $r$ (controlling surface smoothness), and scalar metallic parameter $m$.
        Additionally, normal bump maps encode high-frequency surface perturbations for fine geometric detail. For reconstruction scenarios without predefined UV mappings, we define the local tangent space with the surface normal as up-direction and align the tangent with the world coordinate system~\citep{vainer2024collaborative}.
    \end{subsection}

    \begin{subsection}{Environment Illumination}
        \label{sec:background:illumination}
        Realistic rendering requires modeling the incoming illumination from all directions, typically represented as an environment map $L_\text{env}(\omega)$ that depends only on direction $\omega$.
        Traditional representations using spherical harmonics or spherical Gaussians are limited in capturing high-frequency lighting details like sharp shadows or bright light sources.
        RENI++~\citep{gardner2023reni++} provides a more condensed expressive representation by learning a compact latent space for realistic illumination patterns.
        Environment maps are decoded from latent codes $\mathbf{z} \in \mathbb{R}^{49 \times 3}$ as:
        \begin{equation}
            \label{eq:reni++_decoding}
            L_\text{env}(\omega) = \exp(f_\theta(\mathbf{z}, \gamma(\omega)))
        \end{equation}
        where $f_\theta$ is the pre-trained decoder and $\gamma(\omega)$ provides positional encoding.
        This enables a low dimensional representation perfectly suited for fast feed-forward reconstruction.
    \end{subsection}

    \begin{subsection}{Large Reconstruction Models and Triplane Representations}
        \label{sec:background:lrm}
        Recent advances in feed-forward 3D reconstruction leverage large transformer models trained on extensive 3D datasets.
        Methods like LRM~\citep{hong2023lrm} and TripoSR~\citep{TripoSR2024} demonstrate that direct image-to-3D reconstruction is feasible without per-object optimization.
        
        These approaches typically use triplane representations to efficiently encode 3D information.
        A triplane $\mathbf{T} \in \mathbb{R}^{3 \times C \times H \times W}$ consists of three orthogonal 2D feature planes.
        For any 3D point $\mathbf{p} = (x, y, z)$, features are extracted by projecting onto each plane:
        \begin{equation}
            \label{eq:triplane_projection}
            \mathbf{f}(\mathbf{p}) = \text{concat}(\mathbf{T}_{xy}(x, y), \mathbf{T}_{yz}(y, z), \mathbf{T}_{zx}(z, x))
        \end{equation}
        These concatenated features are then decoded through MLPs to predict geometric and appearance properties.
        SF3D~\citep{sf3d2024} exemplifies this paradigm, it encodes input images with DINOv2~\citep{oquab2023dinov2}, processes them through a transformer with camera conditioning, and outputs triplane features.
        These are decoded into geometry via DMTet~\citep{shen2021dmtet} and textured using fast UV unwrapping.
        However, SF3D is limited to single-view input, global material properties, and lacks environment estimation. Limitations our approach addresses through multi-view fusion and spatially varying material prediction.
    \end{subsection}
    \input{figures/tex/overview}
\end{section}

%% file: figures/tex/overview.tex
\begin{figure*}[tb]
    \centering
\includegraphics[width=\linewidth]{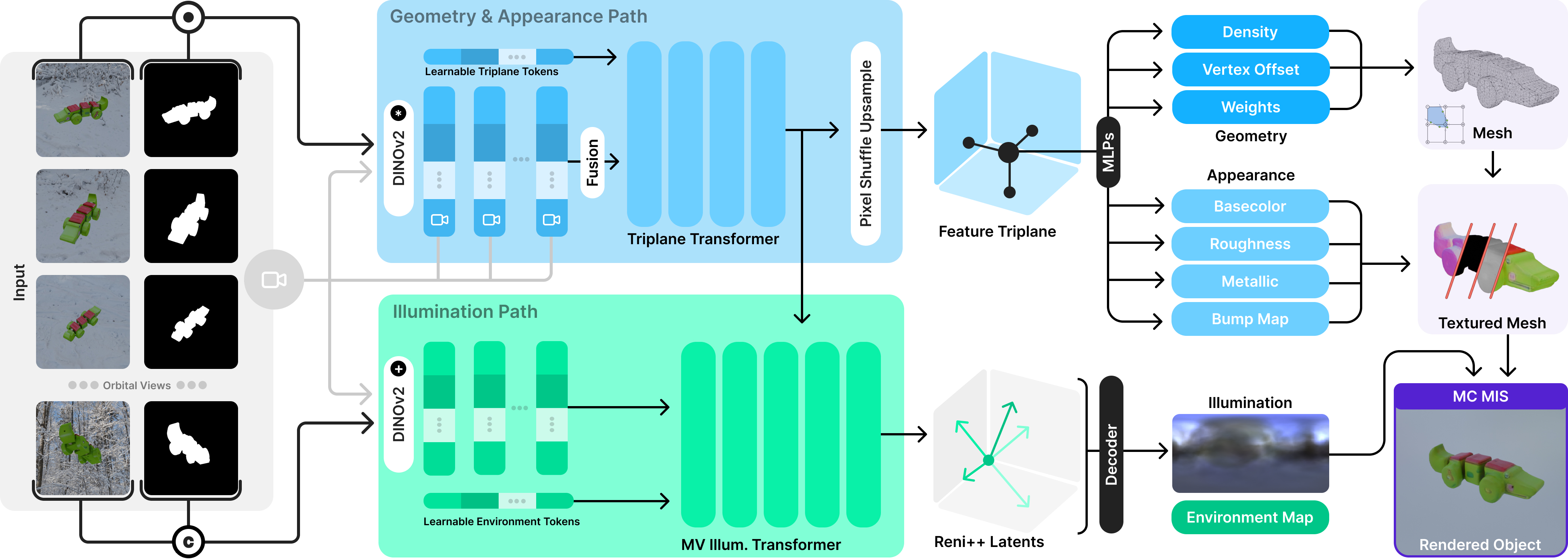}
\titlecaption{\ours Overview}{
Multi-view input images are fused by a shared cross-conditioning transformer into two parallel paths: a Geometry \& Appearance Path (blue) using a Triplane Transformer to predict mesh geometry and PBR materials, and an Illumination Path (green) using a Multi-View Illumination Transformer to estimate HDR environments. Both paths are unified through a differentiable Monte Carlo Multiple Importance Sampling rendering to learn to wproduce complete relightable 3D assets.
}
    \label{fig:overview}
\end{figure*}

%% file: content/04_method.tex
\begin{section}{Method}
\label{sec:method}

Our core insight is that multi-view constraints provide the missing information to disentangle material properties from lighting effects, a problem that remains fundamentally ill-posed for single-view methods. 
We achieve this through a unified two-path architecture that jointly predicts object structure with spatially varying materials and environment illumination from arbitrary numbers of input views.
\Cref{fig:overview} illustrates our complete pipeline.

\begin{subsection}{Multi-view Illumination Disentanglement Architecture}
    \label{sec:method:architecture}
    
    Our approach centers on a novel two-path prediction strategy enabled by multi-view fusion.
    The \textbf{geometry+appearance path} predicts mesh structure and spatially varying BRDF parameters from unified triplane features, while the \textbf{illumination path} estimates HDR environment maps via our multi-view RENI++ extension.
    Both paths are driven by a shared cross-conditioning transformer that fuses arbitrary numbers of input views, creating consistent feature representations that enable robust material-lighting disentanglement.

    \begin{subsubsection}{Cross-view Feature Fusion}
        \label{sec:method:cross_view_fusion}

        Let the input be a set of $N$ masked images with cameras $\{(\mathbf{I}_i,\mathbf{M}_i,\mathbf{C}_i)\}_{i=1}^N$.
        We first form per-view tokens with DINOv2 and camera modulation:
        \begin{equation}
            \mathbf{T}_i^{\text{img}} = \text{DINOv2}(\mathbf{I}_i \odot \mathbf{M}_i),\quad
            \mathbf{e}_i = f_{\text{cam}}(\mathbf{C}_i),\quad
            \mathbf{T}_i^{\text{cond}} = \big[\, \mathbf{T}_i^{\text{img}} \odot \mathbf{e}_i \; ;\; \mathbf{e}_i \,\big].
            \label{eq:cvf_tokens}
        \end{equation}

        We designate one view as the hero view $h$ and its tokens are concatenated to the learned triplane token bank $\mathbf{T}^{\text{tri}}$ and drive the query stream of the transformer:
        \begin{equation}
            \mathbf{Q}_0 = \big[\, \mathbf{T}^{\text{tri}} \;;\; \mathbf{T}_h^{\text{img}} \,\big].
            \label{eq:hero_query}
        \end{equation}
        The hero view serves as the query stream for cross-conditioning and is selected uniformly at random during training and evaluation, ensuring robust performance independent of viewpoint choice.

        To make cross-view context compact yet expressive, we employ latent mixing. 
        A bank of learnable latent tokens $\mathbf{L}_0\!\in\!\mathbb{R}^{L\times D}$ is mixed with the projected cross-view tokens (all non-hero views) to form a memory $\mathbf{M}$ that the query stream will attend to:
        \begin{align}
            \mathbf{H}_i &= P_{\ell}\big(\text{LayerNorm}(\mathbf{T}_i^{\text{cond}})\big), \; i \in \mathcal{V}_{\text{cross}}, \\
            \mathbf{L}_1 &= \text{SelfAttn}(\text{LayerNorm}(\mathbf{L}_0)) \\
            \mathbf{M} &= \text{Interleave}\big(\mathbf{L}_1,\; \text{TokenConcat}(\{\mathbf{H}_i\}_{i\in\mathcal{V}_{\text{cross}}})\big).
            \label{eq:latent_mixing}
        \end{align}
        Here $P_{\ell}$ projects tokens to the latent dimensionality $D$, and Interleave denotes the two-stream interleaved transformer, which alternates blocks that (i) update $\mathbf{Q}$ with cross-attention to $\mathbf{M}$ and (ii) refine $\mathbf{M}$ via self-/cross-attention.
        The main transformer thus computes:
        \begin{equation}
            \mathbf{T}^{\text{out}} = \text{TwoStream}(\mathbf{Q}_0,\, \mathbf{M})\,,
            \label{eq:twostream_out}
        \end{equation}
        which yields triplane-conditioned features that are consistent across an arbitrary number of input views while preserving a dedicated hero view pathway for stable geometry/appearance alignment. 
        In implementation, we use pixel-shuffle upsampling to obtain higher-resolution triplanes from raw predictions.
    \end{subsubsection}

    \begin{subsubsection}{Spatially Varying Material Prediction}
        \label{sec:method:material_prediction}
        
        Our \textbf{geometry+appearance path} operates on the unified triplane representation to predict spatially varying material properties and mesh structure.
        The transformer output tokens $\mathbf{T}^{\text{out}}$ are directly interpreted as triplane pixels, forming our unified 3D representation $\mathbf{T} \in \mathbb{R}^{3 \times 40 \times 384 \times 384}$.
        For any 3D point $\mathbf{p}$, we extract features via triplane projection as established in \Cref{eq:triplane_projection}.
        
        Crucially, we predict all material and geometric properties from this single shared triplane embedding using task-specific MLP heads:
        \begin{equation}
            \{\sigma, \rho, r, m, \mathbf{n}_{\text{bump}}\}(\mathbf{p}) = \{\text{MLP}_{\text{density}}, \text{MLP}_{\text{albedo}}, \text{MLP}_{\text{rough}}, \text{MLP}_{\text{metal}}, \text{MLP}_{\text{normal}}\}(\mathbf{f}(\mathbf{p}))
        \end{equation}
        where $\sigma$ is density, $\rho$ is albedo, $r$ is roughness, $m$ is metallic, and $\mathbf{n}_{\text{bump}}$ represents normal perturbations.
        This unified approach eliminates the need for separate material tokens and enables complex multi-material object support.
        
        Geometry is extracted using Flexicubes~\citep{shen2023flexicubes} for superior mesh quality, and the resulting mesh is textured with spatially varying PBR parameters via fast UV unwrapping.

    \end{subsubsection}

    \begin{subsubsection}{Multi-view Environment Estimation}
        \label{sec:method:environment_estimation}
        
        We introduce a novel multi-view illumination inference approach that fundamentally differs from existing methods. While prior work typically predicts environment maps using simple MLPs from triplane features or single-view observations, we present the first method to leverage multi-view reasoning with adaptive background masking for robust environment estimation.
        
        Our \textbf{illumination path} operates in parallel to the geometry reconstruction, enabling dual-mode operation where our method can robustly recover HDR environments from either direct background observations or indirect material reflectance cues across multiple viewpoints.
        We utilize RENI++ as an efficient illumination representation, however this approach could be easily extended to other lighting representations.

        We encode mask--image pairs $(\mathbf{M}_i,\,\mathbf{I}_i)$ via a trainable DINOv2-small with two extra input channels to obtain mask-aware tokens
        \begin{equation}
            \mathbf{T}_i^{\text{mask}} = f_{\text{mask}}\big([\mathbf{M}_i,\,\mathbf{I}_i]\big),\quad i=1\ldots N.
        \end{equation}
        These tokens are concatenated with the object-transformer outputs to form the environment context
        \begin{equation}
            \mathbf{T}^{\text{env-ctx}} = \text{concat}\big(\{\mathbf{T}_i^{\text{mask}}\}_{i=1}^N,\, \mathbf{T}^{\text{out}}\big).
        \end{equation}
        A dedicated 1D transformer maps learned environment tokens to a RENI++ latent \emph{and} a global rotation (6D) via cross-attention:
        \begin{equation}
            [\,\mathbf{z}_{\text{env}},\, \mathbf{r}_{6\text{D}}\,] = \text{EnvTransformer}\big(\mathbf{T}^{\text{env-bank}},\, \mathbf{T}^{\text{env-ctx}}\big),\quad \mathbf{z}_{\text{env}}\!\in\!\mathbb{R}^{K\times d}, \; \mathbf{r}_{6\text{D}}\!\in\!\mathbb{R}^{6},
        \end{equation}
        where $K\times d$ matches the RENI++ latent grid dimensionality. The final HDR environment is decoded as established in \Cref{eq:reni++_decoding}.
        
        Critically, our training employs stochastic background masking, randomly occluding background pixels in a subset of views during training. This forces the network to solve two complementary tasks: when background pixels are visible, it can read lighting directly from the environment; when they are masked, it must infer lighting from indirect cues in object reflections and shading. This dual mode training enables robust illumination inference in real-world scenes where backgrounds are often partially cropped, saturated, or noisy.
    \end{subsubsection}

\end{subsection}

\begin{subsection}{Disentangled Training via MC+MIS}
    \label{sec:method:training}
    
    Our differentiable physically based Monte Carlo (MC) renderer with Multiple Importance Sampling (MIS) ties both reconstruction paths together, enforcing physically meaningful material-illumination disentanglement while enabling mixed-domain training.
    We found that utilizing VNDF sampling~\citep{Heitz2018GGX} with spherical caps~\citep{Dupuy2023vndfcaps} and antithetic sampling~\citep{zhang2021antithetic} helps stabilize the training.
    This MC+MIS approach enables the following capabilities:

    \begin{itemize}
    \item \textbf{Physical disentanglement}: The renderer enforces that predicted materials $f_r$ and illumination $L_{\text{env}}$ must jointly explain observed images through physically based light transport.
    \item \textbf{Mixed supervision}: When PBR ground truth exists, we additionally use direct material supervision; otherwise, the renderer ensures material and lighting consistency purely through image reconstruction.
    \item \textbf{Domain bridging}: This allows seamless training across synthetic PBR data, synthetic RGB-only renders, and most importantly real-world captures, dramatically improving generalization and robustness.
    \end{itemize}
    
    The result is the first system capable of learning spatially varying material reconstruction from mixed-domain data without supervision collapse, enabling robust performance on real-world inputs while maintaining physical plausibility.
\end{subsection}

\end{section}

%% file: content/05_experiments.tex
\begin{section}{Experiments}
    \label{sec:experiments}
    
    We evaluate \ours across three core dimensions that validate our central thesis: multi-view constraints enable superior material and lighting disentanglement for fast, production ready 3D asset creation.
    Our experiments demonstrate that while we achieve competitive geometry reconstruction at interactive speeds, our primary contribution lies in illumination disentanglement, delivering spatially varying PBR materials and coherent HDR environments that enable high-fidelity relighting.
    
    \begin{subsection}{Implementation and Evaluation Setup}
    \label{sec:experiments:setup}
    We train on 174k objects total: 42k synthetic PBR (full material supervision), 70k synthetic RGB-only, and 62k real-world captures from UCO3D~\citep{liu24uco3d}.
    For evaluation, we test on out-of-distribution datasets including Google Scanned Objects (GSO)~\citep{downs2022google}, Polyhaven~\citep{polyhavenPolyHaven} objects rendered with HDRI-Skies~\citep{HDRISkiesDownload}, Stanford ORB~\citep{kuang2024stanford}, and challenging real-world UCO3D captures with motion blur and imperfect masks.
    We compare against recent feed-forward and generative methods: SF3D~\citep{sf3d2024}, SPAR3D~\citep{huang2025spar3d}, 3DTopia-XL~\citep{chen2024primx}, and Hunyuan3D~\citep{zhao2025hunyuan3d}.
    All experiments run on a single H100 GPU, including mesh extraction and texture baking. 
    To ensure fair comparison, we apply rigid ICP alignment to ground truth meshes before evaluating image metrics, as baselines often produce meshes in arbitrary canonical spaces. \ours predictions are naturally aligned, highlighting a useful feature for practical applications. For more details, please refer to the appendix~\Cref{sec:implementation_details}.
    \end{subsection}

    \input{figures/tex/relighting}

    \begin{subsection}{Material-Lighting Disentanglement: Our Core Contribution}
    \label{sec:experiments:disentanglement}
    
    While overall 3D reconstruction is important, we are particularly interested in the quality of material estimation and illumination disentanglement.

    \textbf{Spatially Varying Material Prediction.} For PBR results in \Cref{fig:pbr_relighting} and \Cref{tab:comparison_pbr}, we demonstrate that \ours predicts fully spatially varying PBR materials that improve significantly with additional views (e.g., where the base of the bed is corrected in \Cref{fig:results_pbr}).
    Our method ranks first across all material metrics: albedo reconstruction achieves 25.00 dB PSNR (vs SF3D's 18.42 dB), roughness reaches 22.69 dB PSNR, and metallic prediction achieves 32.73 dB.
    Multi-view input further enhances these results, demonstrating that cross-view constraints successfully resolve material-lighting ambiguities.

    \textbf{Relighting Performance.} The ultimate test of material-lighting disentanglement is relighting under novel environments.
    For quantitative relighting evaluation, we rendered each reconstruction in a novel out-of-distribution HDR environment.
    Even when competing methods receive ground-truth environment maps as input, \ours ranks first across all relighting metrics in \Cref{tab:comparison_pbr}.
    Visually, \Cref{fig:pbr_relighting} shows that our material estimation is so accurate that the relit reconstructions closely resemble the ground truth, confirming that our material decomposition generalizes well to novel lighting conditions.
    
    \textbf{Environment Estimation.} \Cref{fig:illumination_comparison} compares our predicted HDR environment maps with ground truth.
    Even a single view suffices to recover the correct sky color and sun direction.
    We also show how background information helps recover correct light sources, and utilizing multiple views helps recover correct light directions, even in dark environments.
    In contrast, SPAR3D often predicts over-smoothed, low-contrast maps with no clear light sources.
    \end{subsection}
    
    \input{tables/pbr_reconstruction}
    \input{figures/tex/illumination}

    \begin{subsection}{Overall Reconstruction Quality}
    \label{sec:experiments:reconstruction}
    
    While geometry reconstruction is not our primary focus, \ours achieves competitive results at unprecedented speed.
    Our model achieves quantitative and qualitative state-of-the-art single-view reconstruction results on out of distribution synthetic (GSO, Stanford ORB) and real-world (UCO3D) data in \Cref{tab:comparison_3d}.
    In the multi-view setting, \ours permorms well on geometric and outperforms on all image metrics while running in avg. 0.31s.
    Supplying just four views improves CD by 27\% and pushes the F-score@0.5 to 0.993, showcasing the effectiveness of our multi-view cross-conditioning at virtually unchanged cost.
    Performance saturation beyond 4--8 views stems from coverage saturation: once surface coverage is sufficient, additional random views often provide redundant information rather than new constraints, leading to marginal gains.
    
    \Cref{fig:results_synthetic} offers an end-to-end comparison across all datasets and methods.
    Competing techniques frequently fail or output planar artifacts, while our multi-view fusion reconstructs complete assets, including hidden backsides, with better ground truth lighting and shadowing.
    For real-world captures, \ours remains robust, and our method improves with multi-view input while others do not (e.g., the face of the teddy bear in \Cref{fig:results_synthetic}).
    
    We acknowledge that specialized high-resolution diffusion methods may achieve superior geometric detail through longer optimization.
    However, our contribution lies in the speed-quality trade-off for material-aware reconstruction: we deliver complete, relightable assets in under a second while running 100× faster than generative approaches like Hunyuan3D.
    \end{subsection}
    
    \input{tables/reconstruction}

    \begin{subsection}{Cross-Domain Training Efficiency}
    \label{sec:experiments:efficiency}
    
    Our mixed-domain training protocol enables robust real-world performance~\Cref{fig:results_real_world} despite training on only 174k objects 10-50× less data than recent large-scale methods.
    The key insight is that multi-view constraints provide stronger supervision signals than massive single-view datasets, enabling efficient learning of material-lighting disentanglement.
    
    We evaluate on real-world Stanford ORB dataset~\citep{kuang2024stanford} to demonstrate generalization (\Cref{tab:real_world_stanfordorb}).
    \ours outperforms all baselines across 3D reconstruction, image quality, and material prediction metrics.
    Multi-view input further improves performance.
    \end{subsection}
    
    \input{tables_rebuttal/real_world_stanfordorb}

    \begin{subsection}{Limitations}
    \label{sec:experiments:limitations}
    
    Although rare, failure cases occur where the decomposition fails to disentangle lighting and materials, resulting in baked-in lighting affecting the material maps.
    This seems to occur when environment lighting is not in domain for the RENI++ prior, most notably when multiple very strong light sources are present.
    
    The largest remaining weakness is the relatively limited resolution of the triplane, limiting texture and geometry resolution in practice, also visible in reconstruction examples against Hunyuan3D.
    While we do not claim to have the best geometry prediction, as other methods spend more time with high-quality diffusion processes, we are confident that our illumination disentanglement structure is a contribution that, with sufficient resources, could help larger methods.
    \end{subsection}

\end{section}

%% file: figures/tex/relighting.tex
\begin{figure*}[tb]
    \centering
        \includegraphics[width=1.0\linewidth]{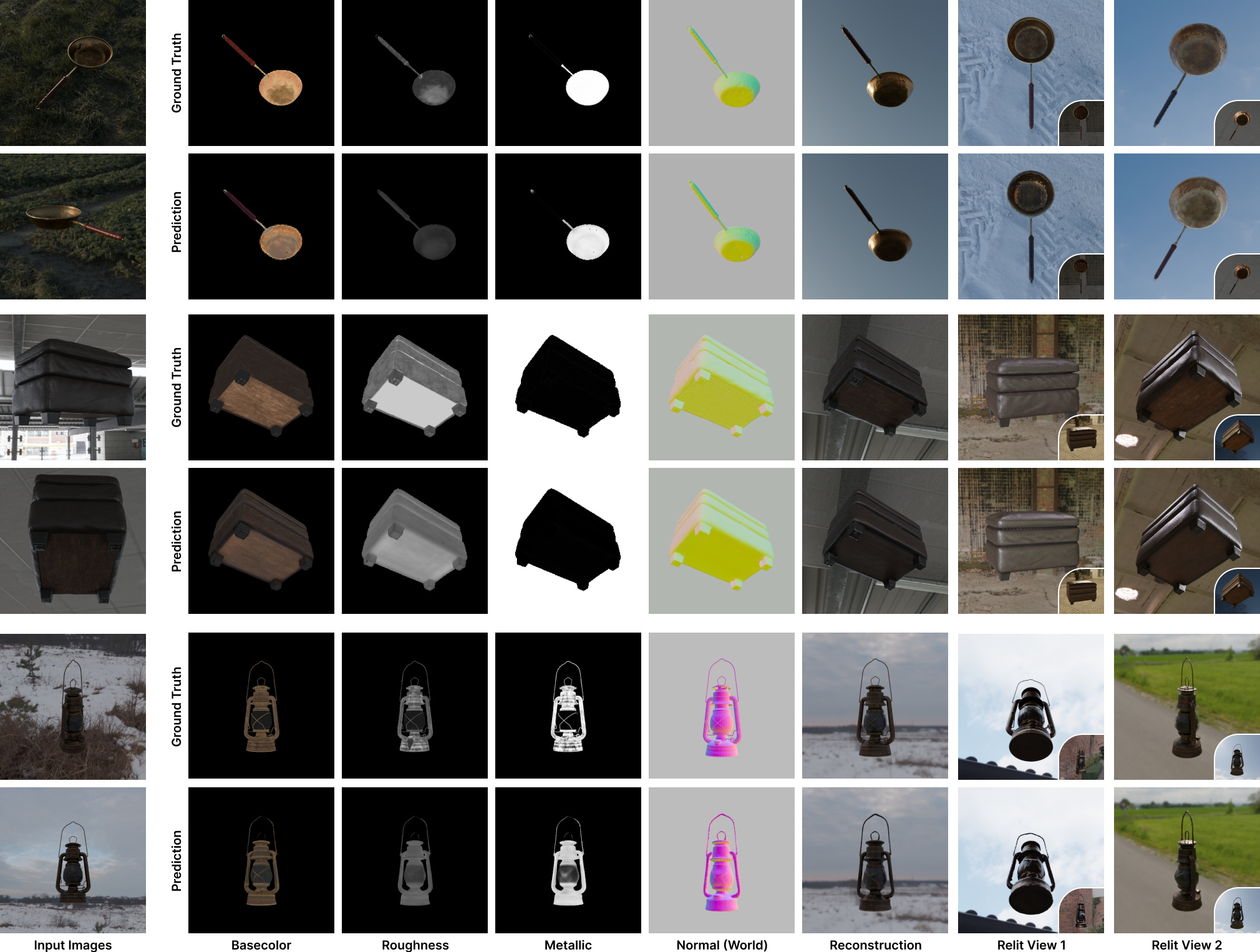}
        \vspace{-0.75cm}
        \titlecaption{PBR \& Relighting Results}{
            We show that our spatially varying PBR prediction is faithful to the ground truth and therefore produces highly detailed and realistic relightings.
        }
    \label{fig:pbr_relighting}
\end{figure*}

%% file: tables/pbr_reconstruction.tex
\begin{table*}[ht]
\centering

\resizebox{0.98\linewidth}{!}{%
\begin{tabular}{lS[table-text-alignment=center]ccccccccccccccc}
 & & \multicolumn{15}{c}{Polyhaven + Blender Shinny} \\
\cmidrule(r){3-17}
& & \multicolumn{3}{c}{Relighting}& \multicolumn{3}{c}{Image}& \multicolumn{3}{c}{Basecolor}& \multicolumn{3}{c}{Roughness}& \multicolumn{3}{c}{Metallic} \\
\cmidrule(r){3-5} \cmidrule(lr){6-8} \cmidrule(lr){9-11} \cmidrule(lr){12-14} \cmidrule(l){15-17}
Method                               & {\textbf{Time (s)}}      & PSNR$\uparrow$                & SSIM$\uparrow$                & LPIPS$\downarrow$             & PSNR$\uparrow$                & SSIM$\uparrow$                & LPIPS$\downarrow$             & PSNR$\uparrow$                & SSIM$\uparrow$               & LSSIMSE$\downarrow$          & PSNR$\uparrow$               & SSIM$\uparrow$               & RMSE$\downarrow$             & PSNR$\uparrow$               & SSIM$\uparrow$               & RMSE$\downarrow$ \\
\midrule
SF3D & \bestoneview{0.26} & \secondbestoneview{15.79} & 0.843 & 0.150 & \secondbestoneview{18.03} & \secondbestoneview{0.875} & 0.120 & 18.42 & 0.831 & \secondbestoneview{0.220} & \secondbestoneview{19.60} & \secondbestoneview{0.876} & 0.127 & 28.37 & 0.888 & 0.116 \\
SPAR3D & 0.36 & 15.23 & 0.836 & 0.154 & 17.02 & 0.862 & 0.132 & 17.70 & 0.822 & 0.251 & 19.53 & 0.874 & \secondbestoneview{0.121} & \secondbestoneview{30.52} & \secondbestoneview{0.895} & 0.088 \\
3DTopia-XL & 31.38 & 14.20 & \secondbestoneview{0.869} & \secondbestoneview{0.140} & 14.60 & 0.853 & 0.168 & 19.52 & 0.818 & 0.330 & 15.16 & 0.847 & 0.191 & 27.60 & 0.861 & \secondbestoneview{0.071} \\
Hunyuan3D & 69.40 & 14.81 & 0.845 & 0.151 & 17.41 & \secondbestoneview{0.875} & \secondbestoneview{0.118} & \secondbestoneview{21.25} & \secondbestoneview{0.837} & 0.265 & --- & --- & --- & --- & --- & --- \\
\textbf{\ours (Ours)} & \secondbestoneview{0.28} & \bestoneview{19.77} & \bestoneview{0.906} & \bestoneview{0.088} & \bestoneview{20.09} & \bestoneview{0.897} & \bestoneview{0.094} & \bestoneview{25.00} & \bestoneview{0.866} & \bestoneview{0.151} & \bestoneview{22.69} & \bestoneview{0.893} & \bestoneview{0.085} & \bestoneview{32.73} & \bestoneview{0.913} & \bestoneview{0.050} \\
\midrule
Hunyuan3D (2 Views) & 41.25 & 14.94 & 0.846 & 0.148 & 17.33 & 0.875 & 0.115 & 21.29 & 0.837 & 0.271 & --- & --- & --- & --- & --- & --- \\
Hunyuan3D (4 Views) & 43.06 & 14.89 & 0.845 & 0.149 & 17.29 & 0.876 & 0.116 & 21.34 & 0.838 & 0.270 & --- & --- & --- & --- & --- & --- \\
\textbf{\ours (Ours)} (2 Views) & \best{0.28} & 20.40 & 0.909 & 0.082 & 21.11 & 0.905 & 0.082 & 25.90 & 0.874 & 0.120 & 23.75 & 0.901 & 0.075 & 33.06 & 0.917 & 0.046 \\
\textbf{\ours (Ours)} (4 Views) & \secondbest{0.29} & 20.94 & 0.912 & 0.078 & 21.48 & 0.909 & 0.078 & 26.45 & 0.878 & 0.112 & 24.08 & 0.904 & 0.072 & 33.18 & \secondbest{0.918} & \secondbest{0.045} \\
\textbf{\ours (Ours)} (8 Views) & 0.31 & \secondbest{21.17} & \secondbest{0.913} & \secondbest{0.076} & \secondbest{21.63} & \secondbest{0.910} & \secondbest{0.076} & \secondbest{26.65} & \secondbest{0.880} & \secondbest{0.111} & \secondbest{24.30} & \secondbest{0.906} & \secondbest{0.071} & \best{33.30} & \best{0.919} & \best{0.044} \\
\textbf{\ours (Ours)} (16 Views) & 0.32 & \best{21.21} & \best{0.914} & \best{0.075} & \best{21.73} & \best{0.911} & \best{0.075} & \best{26.78} & \best{0.881} & \best{0.109} & \best{24.50} & \best{0.907} & \best{0.069} & \secondbest{33.21} & \best{0.919} & \best{0.044} \\
\end{tabular}%
}
\titlecaption{Relighting \& Image \& PBR Metrics Comparison}{
    (Left) Relighting performance. (Middle) Image reconstruction performance. (Right) PBR material reconstruction performance. While most methods produce only global PBR parameters, ours produce spatially varying material maps which increase in quality with more views.
}
\label{tab:comparison_pbr}
\end{table*}

%% file: figures/tex/illumination.tex
\begin{figure*}[tb]
    \centering
        \includegraphics[width=1.0\linewidth]{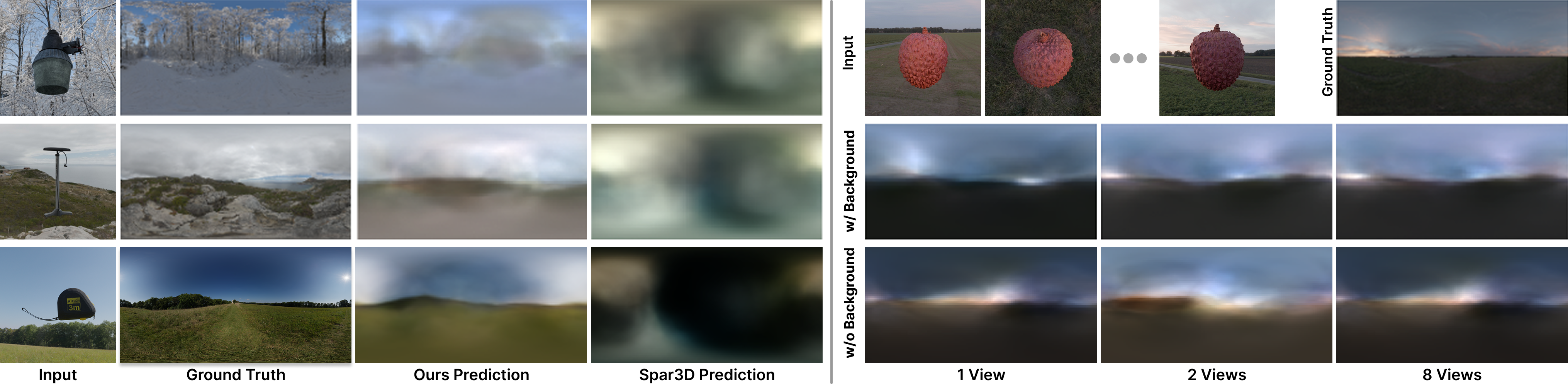}
        \vspace{-0.75cm}

        \titlecaption{Illumination Comparison}{(Left) Single view, illumination prediction results compared to ground truth and SPAR3D, which also predicts RENI++ latents, indicating our severely improved method. (Right) Influence of increasing numbers of views and background information. Notice how well we can predict the illumination in the top rows with background information locate light sources correctly, whereas the bottom row is more spread out as it is inferred from diffuse surface reflections only.
        }
    \label{fig:illumination_comparison}
\end{figure*}

%% file: tables/reconstruction.tex
\begin{table*}[ht!]
\centering

\resizebox{\linewidth}{!}{ %

\begin{tabular}{lS[table-text-alignment=center]cccccccccccccc}
  &  & \multicolumn{7}{c}{Gso + Standford Orb} & \multicolumn{7}{c}{Uco3D} \\
\cmidrule(r){3-9} \cmidrule(l){10-16}
 & & \multicolumn{4}{c}{3D}& \multicolumn{3}{c}{Image}& \multicolumn{4}{c}{3D}& \multicolumn{3}{c}{Image} \\
\cmidrule(r){3-6} \cmidrule(lr){7-9} \cmidrule(lr){10-13} \cmidrule(l){14-16}
Method                            & {\textbf{Time (s)}}      & CD$\downarrow$                 & FS$@$0.1$\uparrow$            & FS$@$0.2$\uparrow$            & FS$@$0.5$\uparrow$            & PSNR$\uparrow$                & SSIM$\uparrow$                & LPIPS$\downarrow$             & CD$\downarrow$                & FS$@$0.1$\uparrow$           & FS$@$0.2$\uparrow$           & FS$@$0.5$\uparrow$           & PSNR$\uparrow$               & SSIM$\uparrow$               & LPIPS$\downarrow$ \\
\midrule
SF3D & \bestoneview{0.28} & \secondbestoneview{0.132} & 0.543 & \secondbestoneview{0.810} & \secondbestoneview{0.974} & \secondbestoneview{17.64} & 0.856 & \secondbestoneview{0.131} & 0.248 & 0.297 & 0.564 & 0.867 & 12.79 & 0.748 & 0.288 \\
SPAR3D & \secondbestoneview{0.39} & 0.152 & 0.507 & 0.766 & 0.959 & 16.34 & 0.837 & 0.151 & 0.232 & \secondbestoneview{0.368} & \secondbestoneview{0.634} & 0.871 & 12.39 & 0.723 & 0.285 \\
TripoSG & 8.54 & 0.232 & 0.357 & 0.619 & 0.881 & 14.47 & 0.832 & 0.211 & 0.274 & 0.297 & 0.520 & 0.842 & 11.85 & 0.752 & 0.330 \\
3DTopia-XL & 45.03 & 0.217 & 0.341 & 0.636 & 0.907 & 14.40 & 0.831 & 0.183 & 0.250 & 0.262 & 0.512 & 0.888 & 12.00 & 0.727 & 0.304 \\
Trellis & 69.09 & 0.149 & \secondbestoneview{0.551} & 0.780 & 0.958 & 16.56 & \secondbestoneview{0.871} & 0.132 & \bestoneview{0.182} & \bestoneview{0.433} & \bestoneview{0.705} & \bestoneview{0.936} & 13.27 & \secondbestoneview{0.760} & 0.309 \\
Hunyuan3D & 39.69 & 0.133 & \bestoneview{0.557} & \bestoneview{0.819} & 0.970 & 16.68 & 0.851 & 0.139 & 0.214 & 0.356 & 0.610 & 0.913 & \secondbestoneview{13.75} & 0.752 & \secondbestoneview{0.273} \\
\textbf{\ours (Ours)} & 0.30 & \bestoneview{0.105} & 0.322 & 0.671 & \bestoneview{0.985} & \bestoneview{19.57} & \bestoneview{0.902} & \bestoneview{0.103} & \secondbestoneview{0.209} & 0.243 & 0.309 & \secondbestoneview{0.935} & \bestoneview{15.28} & \bestoneview{0.839} & \bestoneview{0.214} \\
\midrule
Hunyuan3D (2 Views) & 43.94 & 0.114 & 0.604 & 0.869 & 0.986 & 17.36 & 0.855 & 0.132 & 0.219 & 0.329 & 0.583 & 0.923 & 13.54 & 0.747 & 0.275 \\
Hunyuan3D (4 Views) & 48.06 & 0.110 & 0.636 & 0.875 & 0.986 & 17.40 & 0.856 & 0.130 & 0.222 & 0.341 & 0.600 & 0.904 & 13.58 & 0.749 & 0.277 \\
\textbf{\ours (Ours)} (2 Views) & 0.31 & 0.088 & 0.752 & 0.914 & 0.991 & 20.72 & 0.885 & 0.090 & 0.190 & 0.343 & 0.611 & 0.952 & 15.45 & \best{0.841} & 0.217 \\
\textbf{\ours (Ours)} (4 Views) & \best{0.28} & \secondbest{0.081} & 0.787 & 0.926 & \secondbest{0.993} & 21.43 & 0.894 & 0.080 & 0.188 & 0.346 & 0.622 & 0.953 & \secondbest{15.60} & \secondbest{0.839} & \secondbest{0.212} \\
\textbf{\ours (Ours)} (8 Views) & \secondbest{0.29} & \best{0.076} & \secondbest{0.815} & \best{0.937} & \best{0.994} & \secondbest{22.14} & \secondbest{0.899} & \secondbest{0.072} & \secondbest{0.186} & \secondbest{0.355} & \secondbest{0.625} & \secondbest{0.954} & 15.48 & 0.838 & 0.219 \\
\textbf{\ours (Ours)} (16 Views) & 0.36 & \best{0.076} & \best{0.817} & \secondbest{0.936} & \secondbest{0.993} & \best{22.29} & \best{0.901} & \best{0.070} & \best{0.184} & \best{0.363} & \best{0.631} & \best{0.955} & \best{15.73} & \secondbest{0.839} & \best{0.210} \\
\end{tabular}%
}
\titlecaption{3D and Image Metrics}{\ours clearly achieves SOTA in single and sparse multi-view reconstruction while also achieving great speeds. It is worth noting that that TripoSG and Hunyuan3D also produce signficantly higher vertex counts (100k+ vs 4.5k for ours). }
\label{tab:comparison_3d}
\end{table*}

%% file: tables_rebuttal/real_world_stanfordorb.tex
\begin{table*}[ht]
\centering

\resizebox{\textwidth}{!}{%
\begin{tabular}{lccccccccc}
  &   \multicolumn{9}{c}{Stanford ORB} \\
\cmidrule(r){2-10}
 & \multicolumn{4}{c}{3D}& \multicolumn{3}{c}{Image}& \multicolumn{2}{c}{Basecolor} \\
\cmidrule(r){2-5} \cmidrule(lr){6-8} \cmidrule(l){9-10}
Method & CD$\downarrow$ & FS$@$0.1$\uparrow$ & FS$@$0.2$\uparrow$ & FS$@$0.5$\uparrow$ & PSNR$\uparrow$ & SSIM$\uparrow$ & LPIPS$\downarrow$ & PSNR$\uparrow$ & SSIM$\uparrow$ \\
\midrule
SF3D & 0.152 & 0.512 & 0.769 & 0.954 & \secondbestoneview{17.75} & \secondbestoneview{0.891} & 0.111 & 18.52 & 0.865 \\
SPAR3D & 0.165 & 0.488 & 0.751 & 0.940 & 17.10 & 0.886 & 0.113 & 17.80 & 0.857 \\
Trellis & 0.152 & 0.561 & 0.782 & 0.948 & 17.13 & 0.888 & 0.112 & --- & --- \\
Hunyuan3D & \secondbestoneview{0.141} & \secondbestoneview{0.571} & \secondbestoneview{0.801} & \secondbestoneview{0.960} & 16.96 & 0.877 & \secondbestoneview{0.110} & \secondbestoneview{21.37} & \secondbestoneview{0.872} \\
\textbf{\ours (Ours)} & \bestoneview{0.116} & \bestoneview{0.608} & \bestoneview{0.856} & \bestoneview{0.980} & \bestoneview{18.68} & \bestoneview{0.907} & \bestoneview{0.098} & \bestoneview{24.21} & \bestoneview{0.891} \\
\midrule
Hunyuan3D (2 Views) & 0.134 & 0.588 & 0.809 & 0.967 & 16.91 & 0.876 & 0.108 & 21.42 & 0.872 \\
Hunyuan3D (4 Views) & 0.136 & 0.579 & 0.810 & 0.966 & 16.83 & 0.877 & 0.108 & 21.46 & 0.873 \\
\textbf{\ours (Ours)} (2 Views) & 0.104 & 0.654 & 0.888 & 0.986 & 19.74 & 0.913 & 0.089 & 25.01 & 0.896 \\
\textbf{\ours (Ours)} (4 Views) & \secondbest{0.094} & 0.718 & \secondbest{0.906} & 0.989 & 20.84 & \secondbest{0.919} & \secondbest{0.082} & 25.33 & 0.900 \\
\textbf{\ours (Ours)} (8 Views) & \best{0.089} & \secondbest{0.745} & \best{0.914} & \best{0.991} & \secondbest{21.21} & \best{0.921} & \best{0.080} & \secondbest{25.50} & \secondbest{0.901} \\
\textbf{\ours (Ours)} (16 Views) & \best{0.089} & \best{0.749} & \best{0.914} & \secondbest{0.990} & \best{21.29} & \best{0.921} & \best{0.080} & \best{25.58} & \best{0.902} \\
\end{tabular}%
}
\vspace{0.1cm}
\titlecaption{Real-world Evaluation on Stanford ORB}{
    Quantitative evaluation on Stanford ORB dataset showing 3D reconstruction, image quality, and basecolor material prediction performance. Our method outperforms baselines across all metrics and improves with more input views.
}
\label{tab:real_world_stanfordorb}
\end{table*}

%% file: content/06_conclusion.tex
\begin{section}{Conclusion}
    \label{sec:conclusion}
    We have enhanced the fundamental challenge of illumination disentanglement in feed-forward 3D reconstruction, enabling the first method to jointly predict spatially-varying PBR materials and coherent HDR environments from sparse image inputs.
    Through our novel two-path architecture and differentiable Monte Carlo training, we demonstrate that proper material-lighting separation is achievable at interactive speeds, delivering production-quality relightable assets in under one second.
    
    This development in illumination disentanglement opens exciting avenues for future research and applications.
    The ability to rapidly generate physically accurate 3D assets from casual captures could transform content creation workflows, enabling real-time asset digitization.
    More broadly, our disentanglement framework could extend beyond reconstruction to enable in-the-wild material understanding; imagine training on objects captured under varying real-world illumination to learn material priors that generalize across lighting conditions.
    
    We release all code, pretrained weights, and dataset generation scripts to accelerate adoption and enable the community to build upon this foundation for the next generation of 3D-aware vision systems.
\end{section}

%% file: content/xx_supplements.tex
\newpage

\section*{Appendix}
\addcontentsline{toc}{section}{Appendix}

This appendix provides technical details and additional experimental validation for our multi-view illumination disentanglement approach. We organize the material as follows: \Cref{sec:further_experiments} presents extended experimental results including detailed PBR comparisons, real-world and synthetic reconstruction examples, and an ablation study that validate our architectural choices. \Cref{sec:implementation_details} offers implementation specifics including loss formulations for mixed-domain training, the progressive training protocol that bridges volumetric and mesh-based rendering. Finally, \Cref{sec:datasets} details our curated training data composition, covering both synthetic dataset construction with full PBR supervision and the extensive preprocessing pipeline required to integrate challenging real-world UCO3D captures for robust domain generalization.

\section{Further Experiments}
\label{sec:further_experiments}

This section extends our experimental validation of our multi-view illumination disentanglement approach, including detailed visual analysis of PBR decomposition quality, comprehensive reconstruction comparisons across synthetic and real-world datasets, and critical ablation studies that validate our architectural design choices.

\subsection{Comparison}

\Cref{fig:results_pbr} demonstrates the superior quality of our spatially varying material predictions compared to existing methods. Unlike previous approaches that predict global material properties or fail to achieve proper material-lighting separation, our method produces detailed albedo, roughness, and metallic maps that exhibit realistic spatial variation.
Particularly noteworthy is our method's ability to handle mixed-material objects.
\input{figures/tex/pbr}

Our method's generalization capabilities are extensively validated across diverse synthetic and real-world scenarios. \Cref{fig:results_synthetic} showcases reconstruction quality on  synthetic objects. 
\input{figures/tex/results_synthetic}
\Cref{fig:results_real_world} provides validation on real-world captures, where imperfect masks, camera estimation errors, and challenging lighting conditions test the robustness of our approach. 
\input{figures/tex/results_real_world}

\paragraph{Real-world Material Prediction}
We demonstrate real-world performance on challenging UCO3D captures with motion blur and cluttered backgrounds (\Cref{fig:material_real_world}).
These examples show the benefit of our multi-view setting (e.g., recovering the front of objects given additional views) and improved material prediction as lighting aligns with ground truth.
Our method successfully separates metallic and non-metallic materials even in challenging real-world settings with strong reflections and blur.

\input{figures_rebuttal/tex/material_real_world}

\paragraph{Complex Multi-material Objects}
We evaluate on complex, multi-material objects from the Blender Shiny dataset (\Cref{fig:varying_materials}), demonstrating that our spatially varying PBR prediction generalizes to complex geometries and real materials.
The figure shows predicted basecolor, roughness, metallic, and normal maps, along with relit renderings in novel environments, confirming robust material decomposition across diverse object types.

\input{figures_rebuttal/tex/varying_materials}

\paragraph{Illumination Disentanglement Quality}
\Cref{fig:illumination_comparison_rebuttal} provides detailed qualitative comparison of illumination prediction results between DiffusionLight, SPAR3D, and our method (ReLi3D).
While DiffusionLight hallucinates completely different environments (e.g., predicting indoor scenes for outdoor inputs), and SPAR3D fails to recover meaningful illumination, ReLi3D accurately mimics the ground truth shape and color of the environment maps.
This demonstrates the effectiveness of our dedicated illumination branch and multi-view reasoning for robust environment estimation.

\input{figures_rebuttal/tex/illumination}

\paragraph{Quantitative Evaluation of Illumination Disentanglement}
\Cref{tab:illumination_quantitative} provides quantitative results comparing ReLi3D, SPAR3D, and a DiffusionLight~\cite{phongthawee2023diffusionlight} baseline on the Polyhaven+HDRI dataset.
ReLi3D achieves comparable relighting PSNR to DiffusionLight (20.88 vs 20.93 dB) while being significantly faster (0.34s vs 21.46s) and supporting multi-view input.
SPAR3D achieves similar speed but significantly lower quality (17.10 dB PSNR), confirming the importance of our dedicated illumination branch architecture.

\input{tables_rebuttal/runtime_psnr}

\subsection{Ablation}
\label{sec:further_experiments:ablation}

\Cref{tab:ablation} shows validation of our key architectural choices, with particular emphasis on the critical role of Monte Carlo rendering in achieving high-quality material-lighting disentanglement. The ablation reveals that removing the Monte Carlo renderer (- MC-Render) significantly degrades image reconstruction quality (19.92 → 17.54 dB PSNR).
This finding underscores a crucial insight: the Monte Carlo renderer with Multiple Importance Sampling is not merely an optimization detail but a fundamental component that enables proper physical disentanglement. 

\input{tables/ablation}

\paragraph{Training Stage Contributions}
Our progressive training pipeline transitions from volumetric rendering through spherical Gaussian approximation stages (128 → 256 → 512 Gaussians) to full Monte Carlo integration, as detailed in \Cref{sec:implementation:training}.
\Cref{tab:stage_contributions} reports the share of the total improvement (Phase~1 → Full MC) contributed by each intermediate stage.
The Gaussian stages with larger batch sizes explain 70--80\% of the 3D coverage gains (CD/FS), confirming they mainly stabilize geometry before expensive rendering.
The Monte Carlo stage accounts for the majority of remaining improvements in material disentanglement (basecolor, roughness, metallic).
The 512-Gaussian stage provides the sweet spot for geometry+runtime, while the final MC finetuning sharpens material maps and relighting without regressing 3D accuracy.

\input{tables_rebuttal/resolution_comparison}

\section{Implementation Details}
\label{sec:implementation_details}

This section provides comprehensive implementation details for our multi-view illumination disentanglement architecture, including loss formulations, training protocols, architectural design choices.

\subsection{Loss Functions}
\label{sec:implementation:losses}

Our training objective combines physically-based image reconstruction with material and illumination supervision, designed to handle mixed-domain datasets with varying levels of ground truth availability.

\paragraph{Image Reconstruction Loss}
The primary training signal compares rendered reconstructions against ground truth images not used as input:
\begin{equation}
    \mathcal{L}_{\text{img}} = 10.0 \mathcal{L}_{\text{MSE,im}} + 2.0 \mathcal{L}_{\text{LPIPS,im}}
\end{equation}
This combination ensures both pixel-level accuracy and perceptual quality.

\paragraph{Geometry and Mask Supervision}
During volumetric training stages, we employ mask binary cross-entropy loss $10.0\mathcal{L}_{\text{mask}}$ for foreground segmentation. Geometry losses $\mathcal{L}_{\text{geom}}$ follow the Flexicubes implementation and weighting scheme for robust mesh extraction.

\paragraph{Material Property Supervision}
Given the mixed nature of our training data, material supervision adapts to ground truth availability:
\begin{equation}
    \mathcal{L}_{\text{mat}} = 10.0 \mathcal{L}_{\text{MSE,PBR}} + 4.0 \mathcal{L}_{\text{cos,nrm}} + 0.05 \mathcal{L}_{\text{flat}}
\end{equation}
where basecolor, roughness, and metallic parameters use MSE supervision when available, surface normals employ cosine similarity loss, and bump maps are regularized toward flatness using local normal direction $\mathbf{n}_{\text{up}} = (0, 0, 1)^T$.

\paragraph{Environment Supervision}
Direct RENI++ latent supervision provides illumination guidance:
\begin{equation}
    \mathcal{L}_{\text{env}} = 0.1 \mathcal{L}_{\text{MSE,RENI}} + 0.02 \mathcal{L}_{\text{demod}}
\end{equation}
When RENI++ ground truth is unavailable, demodulation regularization biases the environment toward neutral white lighting.

\subsection{Training Protocol}
\label{sec:implementation:training}

Our multi-stage training protocol progressively transitions from volumetric to mesh-based rendering, culminating in full Monte Carlo integration.

\paragraph{Multi-stage Rendering Pipeline}
We execute three distinct training phases:
\begin{enumerate}
    \item \textbf{Volumetric rendering} of the implicit field using NeRFAcc for initial shape learning
    \item \textbf{Mesh rendering with spherical Gaussian approximation}, progressively increasing image resolution (128 → 256 → 512) for efficient lighting approximation
    \item \textbf{Full Monte Carlo integration} with VNDF sampling, spherical caps, and antithetic sampling for physically accurate shading
\end{enumerate}

Each stage spans 60,000 training steps. This progressive approach ensures stable convergence while gradually increasing rendering fidelity.

\paragraph{Stage-specific Losses and Training}
All stages employ the same loss formulation combining image reconstruction ($\mathcal{L}_{\text{img}}$), material supervision ($\mathcal{L}_{\text{mat}}$), geometry regularization ($\mathcal{L}_{\text{geom}}$), and environment supervision ($\mathcal{L}_{\text{env}}$) as detailed in \Cref{sec:implementation:losses}. Stages 1-3 use spherical Gaussian approximation for lighting, while stage 4 employs full Monte Carlo integration. All network components remain trainable throughout all stages no modules are frozen. The background module is excluded from weight loading when transitioning between stages to allow adaptation to new rendering configurations.

\paragraph{Training Configuration}
We utilize $512 \times 512$ input resolution and randomly sample 1--4 conditioning views per training iteration. The entire pipeline trains end-to-end with a learning rate of $5 \times 10^{-5}$. Batch sizes adapt to computational demands: 64 during volumetric rendering, 192 during spherical Gaussian stages, and 32 during Monte Carlo integration.

\subsection{Architectural Design Choices}
\label{sec:implementation:architecture}

\paragraph{Hero View Selection and Sensitivity}
\label{sec:implementation:hero_view}
The hero view serves as the query stream for the cross-conditioning transformer, providing a stable reference for geometry and appearance alignment.
In our reported metrics (Tables~1 and 2), the hero view is selected uniformly at random, ensuring our results reflect robust performance independent of viewpoint choice, unlike methods relying on canonical frontal views.
To test sensitivity, we compared random selection against fixed frontal-view selection (\Cref{tab:hero_view_sensitivity}).
Results show only marginal differences, with slight perceptual gains for random views likely due to parallax information from side views.

\input{tables_rebuttal/frontal_view}

\paragraph{Illumination Prior and Alternative Representations}
Our framework is compatible with alternative lighting representations: we use spherical Gaussian approximations in the intermediate training stage (\Cref{sec:implementation:training}) before switching to Monte Carlo rendering with RENI++ envmaps.
In those stages, we train with a low frequency Gaussian representation and observe that it fails to capture sharp highlights and directional suns, leading to worse relighting metrics as shown in Table~3.

RENI++ provides a compact but high-frequency representation critical for photorealistic relighting and accurate material and lighting separation.
While nothing in our architecture prevents using SH or Gaussians, we found RENI++ to be the best trade-off between expressiveness and efficiency.
We choose this compact representation to fit our memory limitations; expanding into a more memory intensive representation (e.g., ENV Map HDR prediction) would not be possible with our constraints.

\section{Datasets}
\label{sec:datasets}

Our training leverages a carefully curated mix of synthetic and real-world data to achieve robust generalization while maintaining physical plausibility. This mixed-domain approach enables learning from both controlled synthetic environments with full material supervision and challenging real-world captures that provide crucial domain adaptation.

\subsection{Synthetic Data Composition}
\label{sec:datasets:synthetic}

Following established protocols while extending coverage, we combine multiple synthetic datasets to maximize training diversity. Our synthetic corpus extends the TripoSR dataset protocol with Amazon Berkeley Objects (ABO)~\cite{collins2022abo} and ARIA~\cite{Pan2023aria}, providing comprehensive material and geometric variation.

\paragraph{Rendering Protocol}
Each object is rendered under three distinct illumination environments, randomly rotated around the vertical axis to prevent lighting bias. Camera focal lengths are sampled from a scaled normal distribution between 22° and 37° to match real-world capture conditions. Objects are normalized to unit scale and centered, with cameras positioned to fill the frame with appropriate padding, followed by slight positional augmentation.

We render significantly more views for objects with PBR ground truth (100 images) compared to RGB-only objects (30 images), providing richer supervision where material information is available. This asymmetric sampling strategy maximizes learning efficiency while accommodating varying supervision levels.

\paragraph{Illumination Environments}
Our synthetic rendering employs 1000 HDRI environments sourced from iHDRI~\cite{HDRISkiesDownload} and Polyhaven~\cite{polyhavenPolyHaven} datasets. These environments are preprocessed to extract RENI++ latent codes, enabling direct illumination supervision during training. This diverse illumination set ensures robust material-lighting disentanglement across varied lighting conditions.

\input{figures_rebuttal/tex/failure_cases}

\subsection{Real-world Data Preparation}
\label{sec:datasets:realworld}

The unCommon Objects in 3D (UCO3D)~\cite{liu24uco3d} dataset provides real-world training data, but requires extensive preprocessing to achieve training compatibility with our synthetic data pipeline.

\paragraph{Quality Filtering}
UCO3D contains numerous challenging samples including motion blur, inaccurate masks, and poor camera estimates. We apply strict quality filtering based on reconstruction and camera estimation scores provided by the dataset's Gaussian Splatting optimization, retaining only objects with scores $\geq 1.0$. This filtering dramatically reduces the dataset size but ensures training stability and prevents degraded supervision signals.

\paragraph{Data Preprocessing Pipeline}
Our preprocessing pipeline, illustrated in, %
applies several critical transformations:

\begin{enumerate}
    \item \textbf{Square cropping and centering}: Objects are consistently cropped to square aspect ratios and centered within frames
    \item \textbf{Intrinsic calibration}: Camera intrinsics are carefully adjusted to account for cropping transformations
    \item \textbf{Valid region tracking}: Due to square cropping, we maintain masks for valid view regions and foreground objects
    \item \textbf{Surface normal estimation}: Monocular normal estimation provides additional geometric supervision
    \item \textbf{Scale normalization}: Scene boundaries are rescaled to align with synthetic example scales
\end{enumerate}

This comprehensive preprocessing ensures seamless integration with synthetic training data while preserving the challenging real-world characteristics that drive domain generalization.

\paragraph{Training Integration}
The processed UCO3D data provides RGB-only supervision without material or illumination ground truth. Our mixed-domain training protocol accommodates this through image-space reconstruction losses while synthetic data provides direct material supervision. This combination enables robust real-world generalization while maintaining physical material properties learned from synthetic supervision.

\section{Limitations and Failure Cases}
\label{sec:limitations}

Although rare, failure cases occur where the decomposition fails to disentangle lighting and materials, resulting in baked-in lighting affecting the material maps (\Cref{fig:failure_cases}).
This primarily occurs when (i) environment lighting falls outside the RENI++ prior distribution, especially with multiple extremely bright, localized light sources, or (ii) strong self-shadowing leads to baked-in lighting in material maps, or (iii) dark scenes make basecolor prediction challenging.
However, even in these challenging cases, \ours still outperforms strong baselines like Hunyuan3D~\cite{zhao2025hunyuan3d}.

The largest remaining weakness is the relatively limited resolution of the triplane ($3\times40\times384\times384$), limiting texture and geometry resolution in practice, also visible in reconstruction examples against Hunyuan3D.
Current blur results primarily stem from this resolution constraint and the DINOv2 fine-tuning bottleneck, not the disentanglement framework itself.

Transparent objects present another limitation: while our density-based NeRF pre-training handles transparency, explicit mesh reconstruction of transparent surfaces remains an open research challenge outside our current scope.

ReLi3D assumes known camera poses and physically plausible materials, which are often violated by generated images from diffusion models.
While single-image inputs generally work well when pose estimation is accurate (e.g., from DUST3R~\cite{wang2024dust3r}), severely bad pose estimation leads to blur artifacts.
Multi-view generations sometimes degrade performance due to pose and appearance inconsistencies, though pairing generated multi-view images with proxy 3D reconstructions could enable adaptation to this regime in future work.

%% file: figures/tex/pbr.tex
\begin{figure*}[h!]
    \centering
    \resizebox{0.98\linewidth}{!}{
        \includegraphics[width=0.9\linewidth]{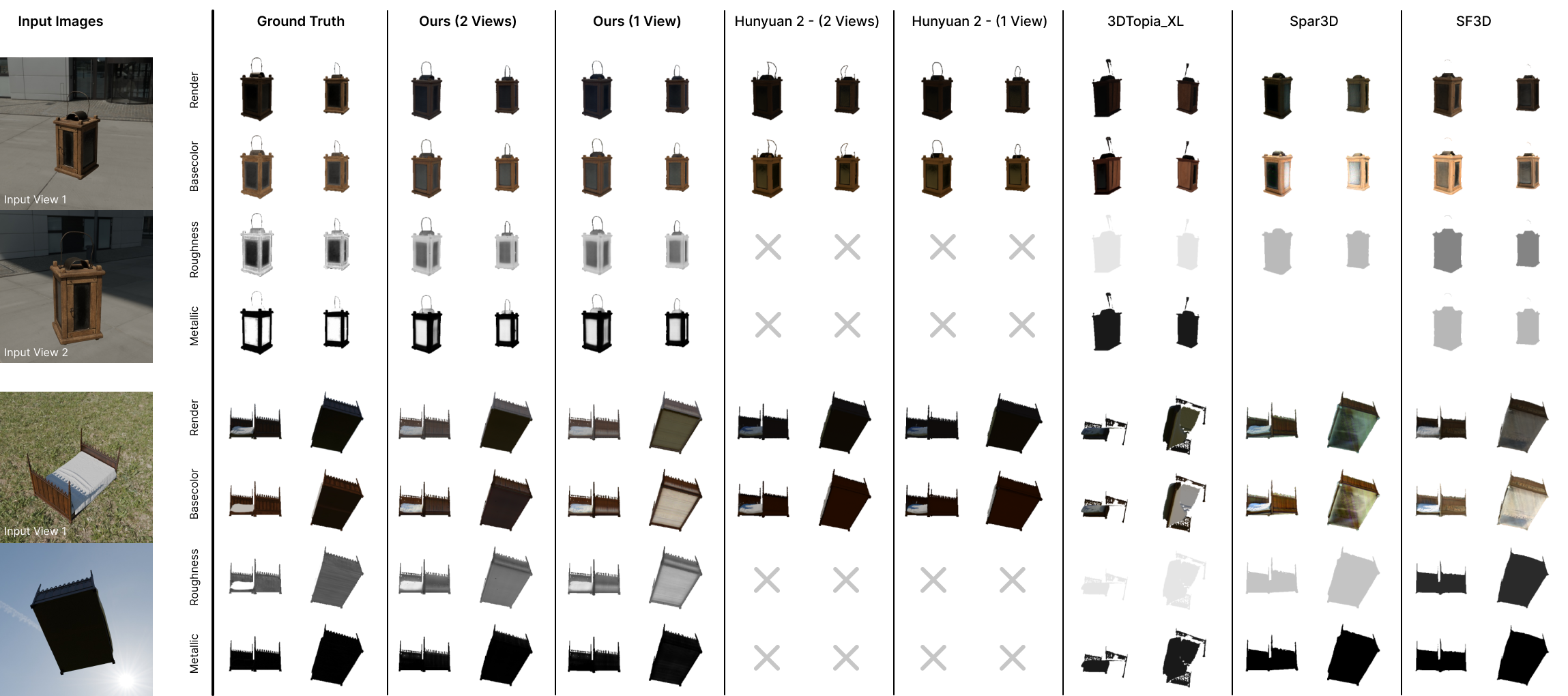}
    }
        \titlecaption{PBR Decomposition Results}{
        Our method is capable of producing highly detailed textures and geometries even from a single view. It is also the only method capable of reproducing accurate spatially varying PBR parameters, which are essential for relighting. 
        }
    \label{fig:results_pbr}
\end{figure*}

%% file: figures/tex/results_synthetic.tex
\begin{figure*}[h!]
    \centering
    \resizebox{0.98\linewidth}{!}{
        \includegraphics[width=0.9\linewidth]{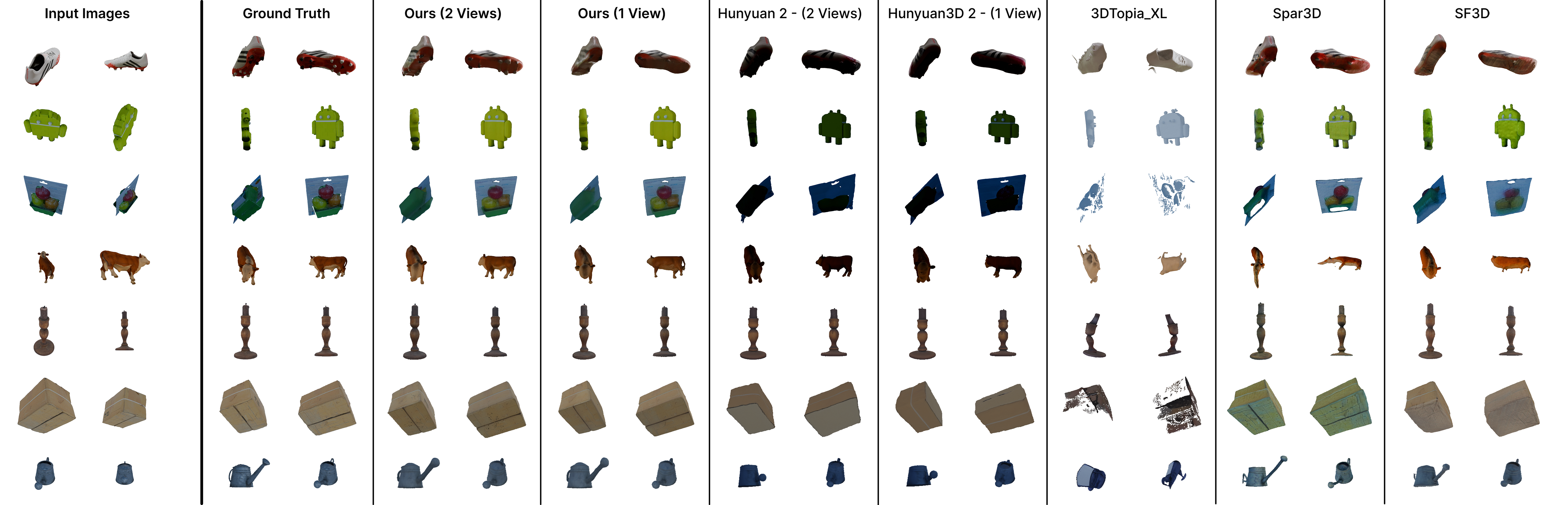}
    }
        \titlecaption{Reconstruction Results (Synthetic)}{
        Our method performs well across synthetic data and shows accurate reconstructions from a single view. Other methods show collaps with bend or flat predictions.  
        }
    \label{fig:results_synthetic}
\end{figure*}

%% file: figures/tex/results_real_world.tex
\begin{figure*}[h!]
    \centering
    \resizebox{0.98\linewidth}{!}{
        \includegraphics[width=0.9\linewidth]{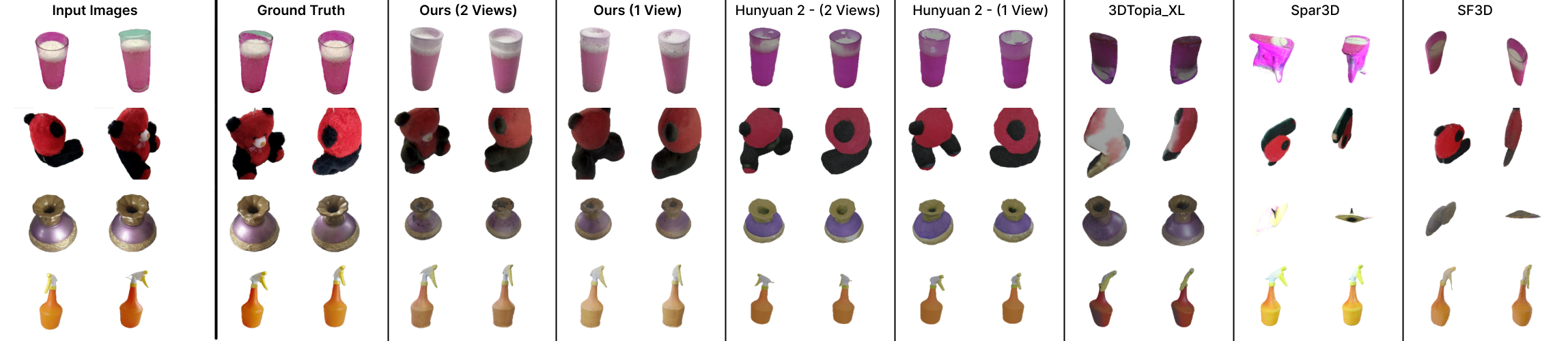}
    }
        \titlecaption{Reconstruction Results (Real World)}{
        Our method produces accurate reconstructions for real-world data, although challenging. Incorporating multiple views improves the performance further by clearing up uncertainties in unseen areas. 
        }
    \label{fig:results_real_world}
\end{figure*}

%% file: figures_rebuttal/tex/material_real_world.tex
\begin{figure}[t]
    \centering
    \includegraphics[width=1.0\linewidth]{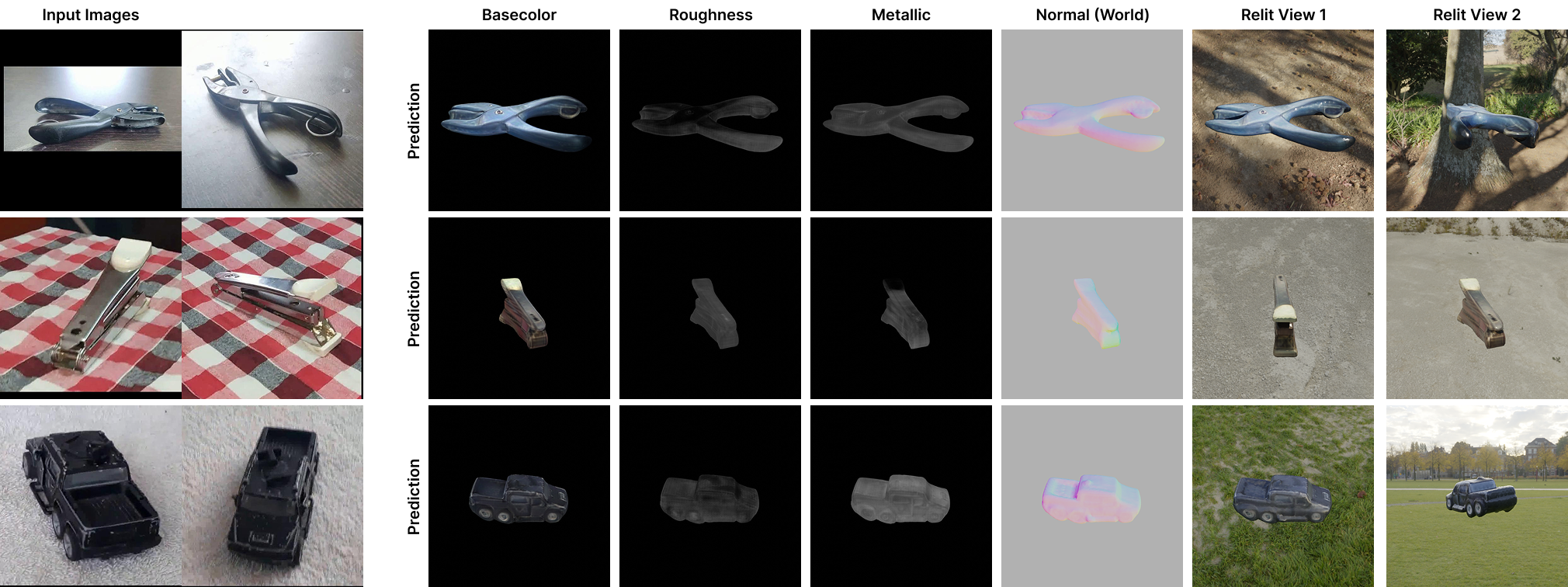}
    \titlecaption{Real-world material prediction}{Material maps (albedo, roughness, metallic, normal) for real-world objects from UCO3D dataset on very challenging settings, strong reflections and blur. Our method is still able to make a rough prediction and faithfully separates metallic and non-metallic materials.}
    \label{fig:material_real_world}
\end{figure}

%% file: figures_rebuttal/tex/varying_materials.tex
\begin{figure*}[h!]
    \centering
    \includegraphics[width=1.0\linewidth]{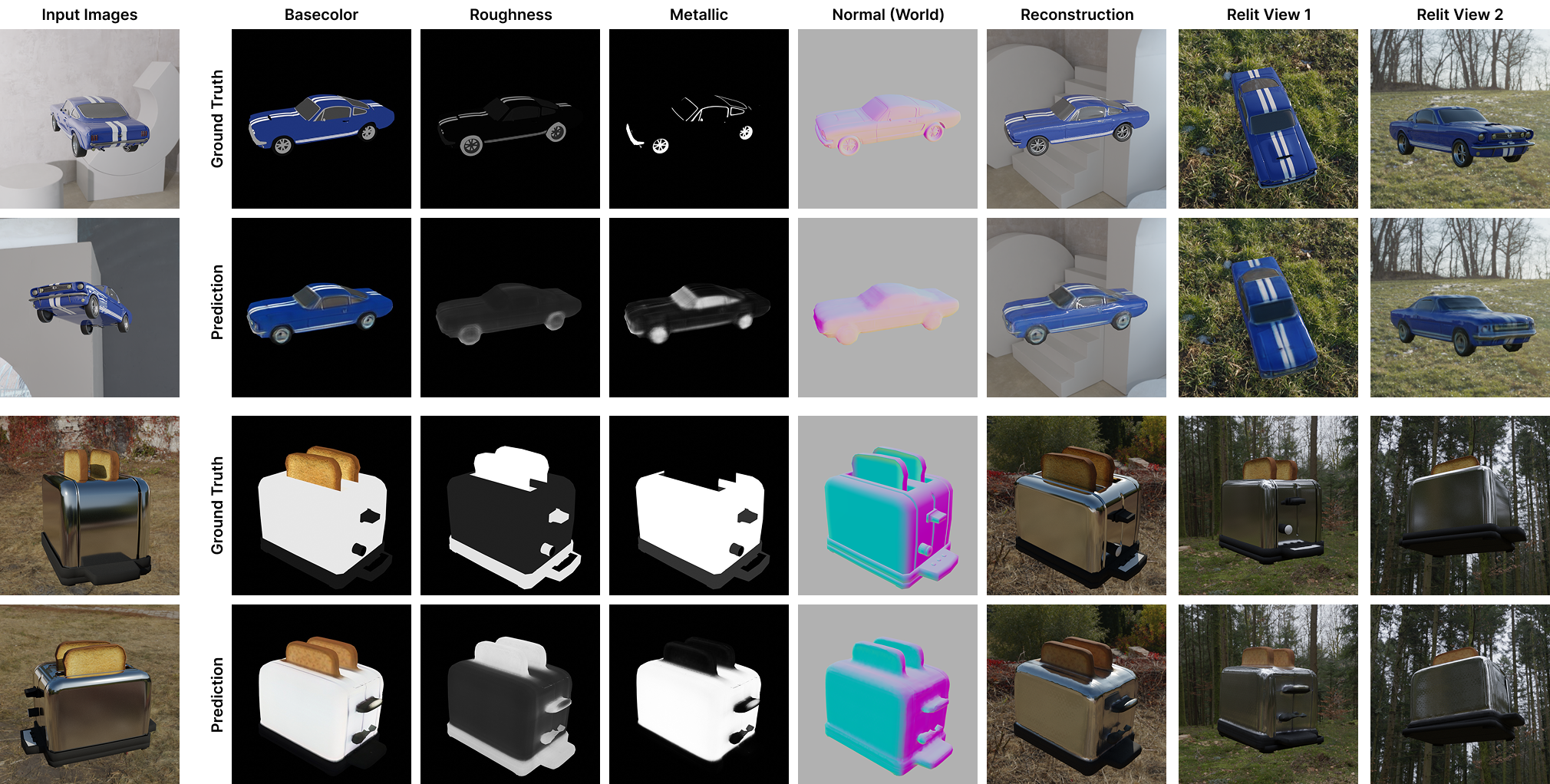}
    \titlecaption{Varying materials and complex objects}{
        Results on the Blender Shiny dataset showing spatially varying PBR material prediction on complex multi-material objects. The figure shows predicted basecolor, roughness, metallic, and normal maps, along with relit renderings in novel environments.
    }
    \label{fig:varying_materials}
\end{figure*}

%% file: figures_rebuttal/tex/illumination.tex
\begin{figure}[t]
    \centering
    \includegraphics[width=1.0\linewidth]{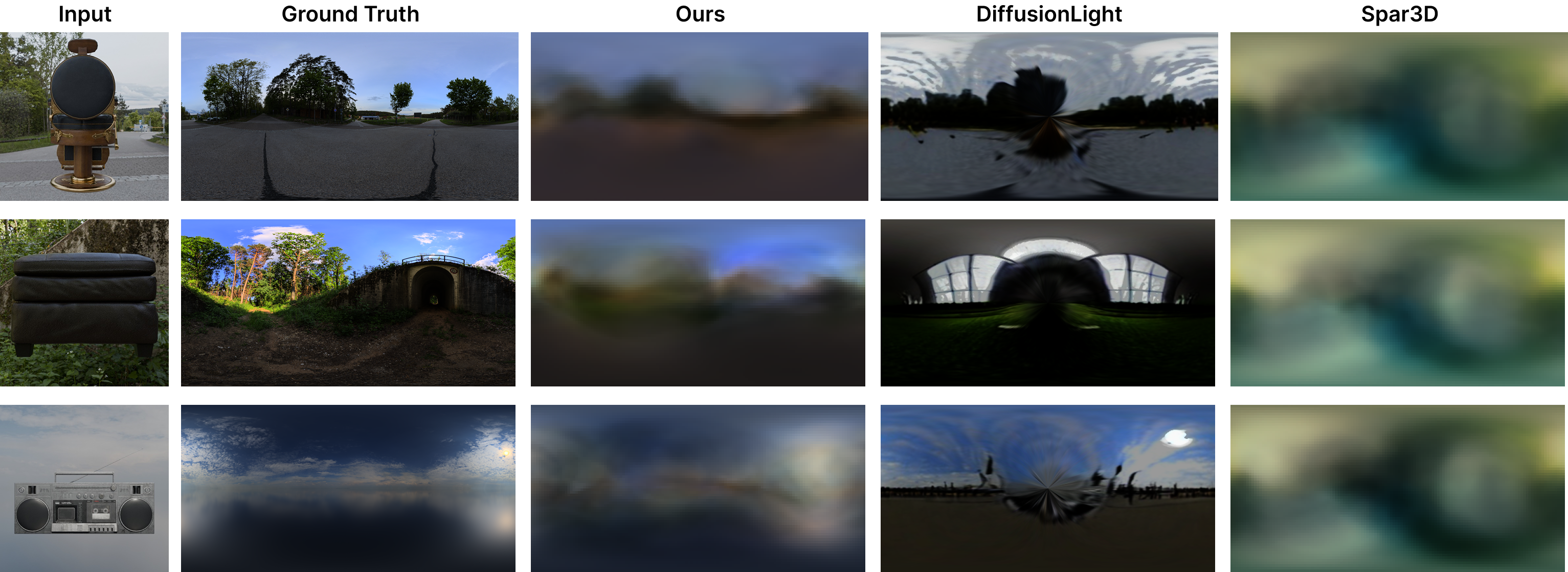}
    \titlecaption{Illumination Comparison}{Comparison of illumination prediction results between DiffusionLight, SPAR3D, and our method (ReLi3D). Predicted environmens vary vastly while ours mimics the ground truth shape and color, DiffusionLight hallucinates a completely different environment, SPAR3D fails.}
    \label{fig:illumination_comparison_rebuttal}
\end{figure}

%% file: tables_rebuttal/runtime_psnr.tex
\begin{table*}[ht]
\centering

\begin{tabular}{ccc|ccc}
\multicolumn{3}{c|}{Time (s)$\downarrow$} & \multicolumn{3}{c}{PSNR$\uparrow$} \\
\cmidrule(r){1-3} \cmidrule(l){4-6}
Diffusion Light & \ours & SPAR3D & Diffusion Light & \ours & SPAR3D \\
\midrule
21.46 & 0.34 & 0.33 & 20.93 & 20.88 & 17.10 \\
\end{tabular}%

\titlecaption{Quantitative evaluation of illumination disentanglement}{Comparison of environment map prediction and relighting quality on Polyhaven+HDRI dataset.}
\label{tab:illumination_quantitative}
\end{table*}

%% file: tables/ablation.tex
\begin{table}[ht!]

\centering

\begin{tabular}{lccccc}
 & \multicolumn{4}{c}{3D} & Image \\
\cmidrule(r){2-5} \cmidrule(r){6-6}
Method & CD$\downarrow$ & FS$@$0.1$\uparrow$ & FS$@$0.2$\uparrow$ & FS$@$0.5$\uparrow$ & PSNR$\uparrow$ \\
\midrule
\ours & \best{0.110} & \best{0.676} & \best{0.870} & \best{0.975} & \best{19.92} \\
 - MC-Render & 0.114 & 0.668 & 0.865 & 0.971 & 17.54

\end{tabular}%

\titlecaption{Ablation study}{Impact of removing the differentiable Monte-Carlo renderer (- MC-Render).}
\label{tab:ablation}
\end{table}

%% file: tables_rebuttal/resolution_comparison.tex
\begin{table*}[ht]
\centering
\resizebox{1.0\linewidth}{!}{%
\begin{tabular}{lccccc}
\toprule
Method & \multicolumn{1}{c}{3D Coverage Share (\%)} & \multicolumn{1}{c}{Image Quality Share (\%)} & \multicolumn{1}{c}{Basecolor Share (\%)} & \multicolumn{1}{c}{Roughness Share (\%)} & \multicolumn{1}{c}{Metallic Share (\%)} \\
\midrule
Phase 2 (256) vs Phase 1 (128) & \secondbest{20.2} & 6.8 & 22.6 & 6.3 & 7.7 \\
Phase 3 (512) vs Phase 2 (256) & \best{70.3} & \best{50.0} & \secondbest{23.9} & \secondbest{31.3} & \secondbest{41.0} \\
MC (Full) vs Phase 3 (512) & 9.5 & \secondbest{43.2} & \best{53.5} & \best{62.4} & \best{51.3} \\
\bottomrule
\end{tabular}%
}
\titlecaption{Training stage contribution analysis}{Average share of the total improvement from Phase~1 (128 Gaussians) to the full Monte Carlo stage that is attributable to each intermediate stage. Columns aggregate the metrics shown in Table~1: (1) 3D coverage (CD and FS@{0.05--0.5}), (2) image quality (PSNR, SSIM, LPIPS), (3) basecolor (PSNR, SSIM, LSSIMSE), (4) roughness (PSNR, SSIM, RMSE), and (5) metallic (PSNR, SSIM, RMSE). Early Gaussian stages mainly expand 3D coverage, while the Monte Carlo refinement sharpens PBR material disentanglement.}\label{tab:stage_contributions}
\end{table*}

%% file: tables_rebuttal/frontal_view.tex
\begin{table*}[ht]
\centering

\resizebox{0.98\linewidth}{!}{%
\begin{tabular}{lccccccc}
  & \multicolumn{7}{c}{Polyhaven Subset} \\
\cmidrule(r){2-8}
 & \multicolumn{4}{c}{3D}& \multicolumn{3}{c}{Image} \\
\cmidrule(r){2-5} \cmidrule(l){6-8}
Method & CD$\downarrow$ & FS$@$0.1$\uparrow$ & FS$@$0.2$\uparrow$ & FS$@$0.5$\uparrow$ & PSNR$\uparrow$ & SSIM$\uparrow$ & LPIPS$\downarrow$ \\
\midrule
\midrule
\textbf{\ours (Random Hero)} & 0.102 & 0.697 & 0.883 & 0.982 & 20.25 & 0.919 & 0.083 \\
\textbf{\ours (Frontal Hero)} & 0.123 & 0.641 & 0.840 & 0.965 & 19.06 & 0.909 & 0.095 \\
\end{tabular}%
}
\titlecaption{Hero view selection sensitivity}{Comparison of metrics using random hero view selection versus always selecting the most frontal view on the Polyhaven dataset.}
\label{tab:hero_view_sensitivity}
\end{table*}

%% file: figures_rebuttal/tex/failure_cases.tex
\begin{figure}[t]
    \centering
    \includegraphics[width=0.95\linewidth]{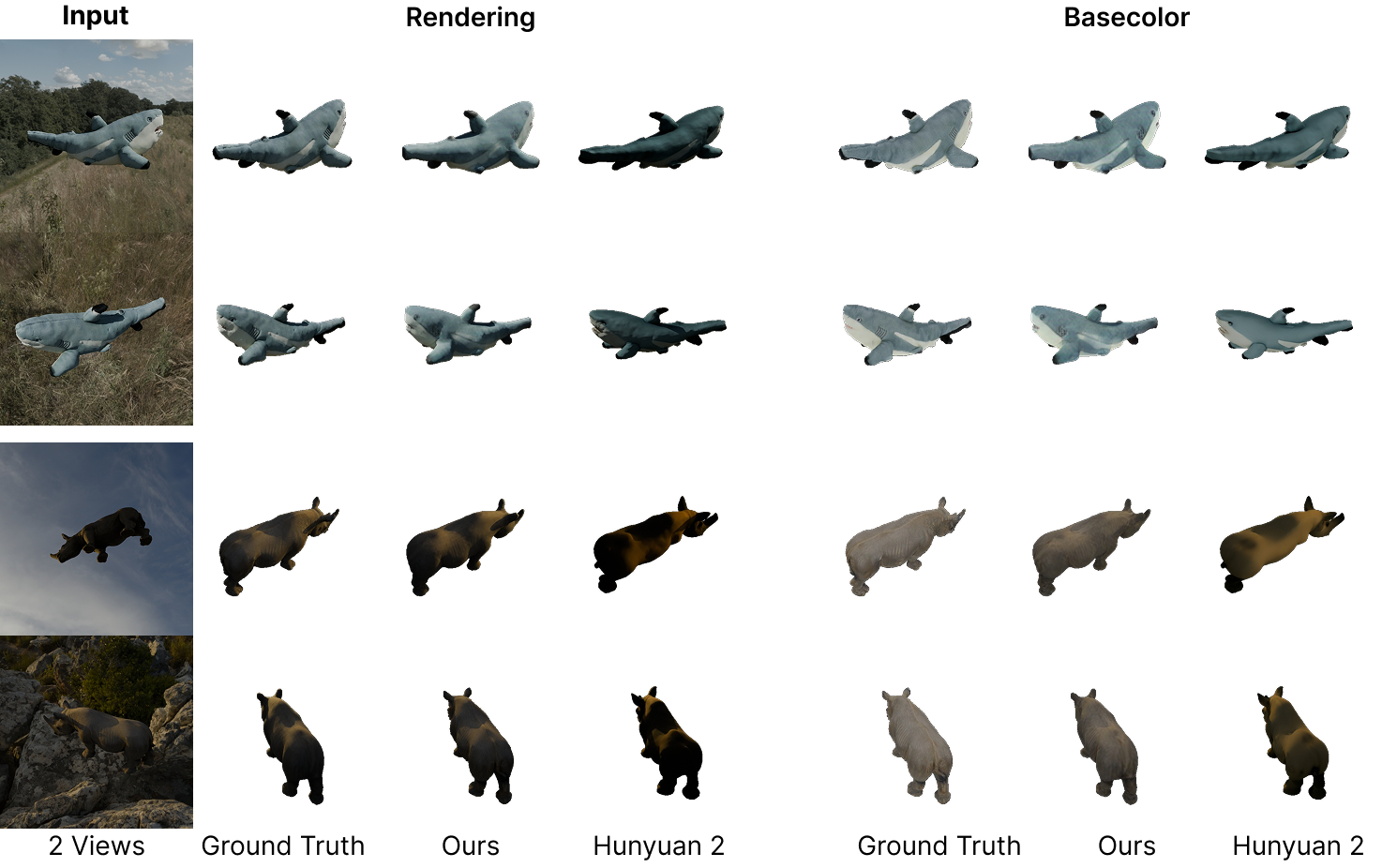}
    \titlecaption{Failure cases}{
        Failure cases showing challenges with baked in lighting for objects with strong self shadowing (fin of shark) and basecolor prediction difficulties in dark scenes (Rhino). Comparison includes results from Hunyuan3D and our method.
    }
    \label{fig:failure_cases}
\end{figure}

%% file: lirm.bib
@String(CVPR= {IEEE Conf. Comput. Vis. Pattern Recog.})

@String(ICCV= {Int. Conf. Comput. Vis.})

@String(ECCV= {Eur. Conf. Comput. Vis.})

@String(TOG= {ACM Trans. Graph.})

@String(ICLR = {Int. Conf. Learn. Represent.})

@String(CVPR  = {CVPR})

@String(ICCV  = {ICCV})

@String(ECCV  = {ECCV})

@String(TOG   = {ACM TOG})

@String(ICLR  = {ICLR})

@article{li2018differentiable,
  title     = {Differentiable monte carlo ray tracing through edge sampling},
  author    = {Li, Tzu-Mao and Aittala, Miika and Durand, Fr{\'e}do and Lehtinen, Jaakko},
  journal   = {ACM Transactions on Graphics (TOG)},
  volume    = {37},
  number    = {6},
  pages     = {1--11},
  year      = {2018},
  publisher = {ACM New York, NY, USA}
}

@inproceedings{boss2021nerd,
  title     = {Nerd: Neural reflectance decomposition from image collections},
  author    = {Boss, Mark and Braun, Raphael and Jampani, Varun and Barron, Jonathan T and Liu, Ce and Lensch, Hendrik},
  booktitle = {Proceedings of the IEEE/CVF International Conference on Computer Vision},
  pages     = {12684--12694},
  year      = {2021}
}

@article{boss2022samurai,
  title   = {Samurai: Shape and material from unconstrained real-world arbitrary image collections},
  author  = {Boss, Mark and Engelhardt, Andreas and Kar, Abhishek and Li, Yuanzhen and Sun, Deqing and Barron, Jonathan and Lensch, Hendrik and Jampani, Varun},
  journal = {Advances in Neural Information Processing Systems},
  volume  = {35},
  pages   = {26389--26403},
  year    = {2022}
}

@article{zhang2021nerfactor,
  title     = {Nerfactor: Neural factorization of shape and reflectance under an unknown illumination},
  author    = {Zhang, Xiuming and Srinivasan, Pratul P and Deng, Boyang and Debevec, Paul and Freeman, William T and Barron, Jonathan T},
  journal   = {ACM Transactions on Graphics (ToG)},
  volume    = {40},
  number    = {6},
  pages     = {1--18},
  year      = {2021},
  publisher = {ACM New York, NY, USA}
}

@inproceedings{engelhardt2024shinobi,
  title     = {SHINOBI: Shape and Illumination using Neural Object Decomposition via BRDF Optimization In-the-wild},
  author    = {Engelhardt, Andreas and Raj, Amit and Boss, Mark and Zhang, Yunzhi and Kar, Abhishek and Li, Yuanzhen and Sun, Deqing and Brualla, Ricardo Martin and Barron, Jonathan T and Lensch, Hendrik and others},
  booktitle = {Proceedings of the IEEE/CVF Conference on Computer Vision and Pattern Recognition},
  pages     = {19636--19646},
  year      = {2024}
}

@inproceedings{liang2024gs,
  title     = {Gs-ir: 3d gaussian splatting for inverse rendering},
  author    = {Liang, Zhihao and Zhang, Qi and Feng, Ying and Shan, Ying and Jia, Kui},
  booktitle = {Proceedings of the IEEE/CVF Conference on Computer Vision and Pattern Recognition},
  pages     = {21644--21653},
  year      = {2024}
}

@inproceedings{barron2013intrinsic,
  title     = {Intrinsic scene properties from a single rgb-d image},
  author    = {Barron, Jonathan T and Malik, Jitendra},
  booktitle = {Proceedings of the IEEE Conference on Computer Vision and Pattern Recognition},
  pages     = {17--24},
  year      = {2013}
}

@inproceedings{li2018materials,
  title     = {Materials for masses: SVBRDF acquisition with a single mobile phone image},
  author    = {Li, Zhengqin and Sunkavalli, Kalyan and Chandraker, Manmohan},
  booktitle = {Proceedings of the European conference on computer vision (ECCV)},
  pages     = {72--87},
  year      = {2018}
}

@article{gardner2017learning,
  title   = {Learning to predict indoor illumination from a single image},
  author  = {Gardner, Marc-Andr{\'e} and Sunkavalli, Kalyan and Yumer, Ersin and Shen, Xiaohui and Gambaretto, Emiliano and Gagn{\'e}, Christian and Lalonde, Jean-Fran{\c{c}}ois},
  journal = {arXiv preprint arXiv:1704.00090},
  year    = {2017}
}

@article{kuang2024stanford,
  title   = {Stanford-ORB: a real-world 3d object inverse rendering benchmark},
  author  = {Kuang, Zhengfei and Zhang, Yunzhi and Yu, Hong-Xing and Agarwala, Samir and Wu, Elliott and Wu, Jiajun and others},
  journal = {Advances in Neural Information Processing Systems},
  volume  = {36},
  year    = {2024}
}


%% file: references.bib
@article{huang2025spar3d,
  author    = {Huang, Zixuan and Boss, Mark and Vasishta, Aaryaman and Rehg, James M and Jampani, Varun},
  title     = {SPAR3D: Stable Point-Aware Reconstruction of 3D Objects from Single Images},
  booktitle = {arXiv preprint arXiv:2501.04689},
  year      = {2025}
}

@article{collins2022abo,
  title   = {ABO: Dataset and Benchmarks for Real-World 3D Object Understanding},
  author  = {Collins, Jasmine and Goel, Shubham and Deng, Kenan and Luthra, Achleshwar and
             Xu, Leon and Gundogdu, Erhan and Zhang, Xi and Yago Vicente, Tomas F and
             Dideriksen, Thomas and Arora, Himanshu and Guillaumin, Matthieu and
             Malik, Jitendra},
  journal = {CVPR},
  year    = {2022}
}

@inproceedings{liu24uco3d,
  author    = {Liu, Xingchen and Tayal, Piyush and Wang, Jianyuan
               and Zarzar, Jesus and Monnier, Tom and Tertikas, Konstantinos
               and Duan, Jiali and Toisoul, Antoine and Zhang, Jason Y.
               and Neverova, Natalia and Vedaldi, Andrea
               and Shapovalov, Roman and Novotny, David},
  booktitle = {arXiv},
  title     = {UnCommon Objects in 3D},
  year      = {2024}
}

@misc{polyhavenPolyHaven,
  author       = {Poly Haven},
  title        = {{P}oly {H}aven • {P}oly {H}aven --- polyhaven.com},
  howpublished = {\url{https://polyhaven.com/}},
  year         = {2024},
  note         = {[Accessed 22-08-2024]}
}

@misc{HDRISkiesDownload,
  title        = {{HDRI Skies - Download your favorite HDRI Sky for Free!}},
  journal      = {iHDRI.COM},
  author       = {IHDRI},
  urldate      = {2024-11-08},
  howpublished = {https://www.ihdri.com/},
  year         = {2024},
  langid       = {ngerman}
}

@inproceedings{Pan2023aria,
  author    = {Pan, Xiaqing and Charron, Nicholas and Yang, Yongqian and Peters, Scott and Whelan, Thomas and Kong, Chen and Parkhi, Omkar and Newcombe, Richard and Ren, Yuheng (Carl)},
  title     = {Aria Digital Twin: A New Benchmark Dataset for Egocentric 3D Machine Perception},
  booktitle = {Proceedings of the IEEE/CVF International Conference on Computer Vision (ICCV)},
  month     = {October},
  year      = {2023},
  pages     = {20133-20143}
}

@article{li2025lirm,
  title   = {LIRM: Large Inverse Rendering Model for Progressive Reconstruction of Shape, Materials and View-dependent Radiance Fields},
  author  = {Li, Zhengqin and Wang, Dilin and Chen, Ka and Lv, Zhaoyang and Nguyen-Phuoc, Thu and Lee, Milim and Huang, Jia-Bin and Xiao, Lei and Zhang, Cheng and Zhu, Yufeng and others},
  journal = {arXiv preprint arXiv:2504.20026},
  year    = {2025}
}

@inproceedings{burley2012disneybrdf,
  title        = {Physically-based shading at disney},
  author       = {Burley, Brent and Studios, Walt Disney Animation},
  booktitle    = {Acm Siggraph},
  volume       = {2012},
  number       = {2012},
  pages        = {1--7},
  year         = {2012},
  organization = {vol. 2012}
}

@article{Heitz2018GGX,
  author  = {Eric Heitz},
  journal = {Journal of Computer Graphics Techniques (JCGT)},
  title   = {Sampling the GGX Distribution of Visible Normals},
  year    = 2018
}

@article{Dupuy2023vndfcaps,
  journal   = {Computer Graphics Forum},
  title     = {{Sampling Visible GGX Normals with Spherical Caps}},
  author    = {Dupuy, Jonathan and Benyoub, Anis},
  year      = {2023},
  publisher = {The Eurographics Association and John Wiley & Sons Ltd.}
}

@article{zhang2021antithetic,
  author  = {Zhang, Cheng and Dong, Zhao and Doggett, Michael and Zhao, Shuang},
  title   = {Antithetic sampling for Monte Carlo differentiable rendering},
  year    = {2021},
  journal = {ACM Trans. Graph.}
}

@article{shen2023flexicubes,
  author     = {Shen, Tianchang and Munkberg, Jacob and Hasselgren, Jon and Yin, Kangxue and Wang, Zian 
                and Chen, Wenzheng and Gojcic, Zan and Fidler, Sanja and Sharp, Nicholas and Gao, Jun},
  title      = {Flexible Isosurface Extraction for Gradient-Based Mesh Optimization},
  year       = {2023},
  issue_date = {August 2023},
  journal    = {ACM Trans. Graph.}
}

@inproceedings{shen2021dmtet,
  title     = {Deep Marching Tetrahedra: a Hybrid Representation for High-Resolution 3D Shape Synthesis},
  author    = {Tianchang Shen and Jun Gao and Kangxue Yin and Ming-Yu Liu and Sanja Fidler},
  year      = {2021},
  booktitle = {Advances in Neural Information Processing Systems (NeurIPS)}
}

@inproceeding{dihmann2024sssgs,
  author    = {Dihlmann, Jan-Niklas and Majumdar, Arjun and Engelhardt, Andreas and Braun, Raphael and Lensch, Hendrik P.A.},
  title     = {Subsurface Scattering for Gaussian Splatting},
  booktitle = {arXiv preprint arXiv:2408.12282},
  year      = {2024}
}

@article{chen2024primx,
  title   = {3DTopia-XL: High-Quality 3D PBR Asset Generation via Primitive Diffusion},
  author  = {Chen, Zhaoxi and Tang, Jiaxiang and Dong, Yuhao and Cao, Ziang and Hong, Fangzhou and Lan, Yushi and Wang, Tengfei and Xie, Haozhe and Wu, Tong and Saito, Shunsuke and Pan, Liang and Lin, Dahua and Liu, Ziwei},
  journal = {arXiv preprint arXiv:2409.12957},
  year    = {2024}
}

@article{wang2024dust3r,
  title   = {DUSt3R: Geometric 3D Vision Made Easy},
  author  = {Wang, Shuzhe and Leroy, Vincent and Cabon, Yohann and Chidlovskii, Boris and Revaud, J{\'e}r{\^o}me},
  journal = {arXiv preprint arXiv:2312.14132},
  year    = {2024}
}

@article{phongthawee2023diffusionlight,
  title   = {DiffusionLight: Light Probes for Free by Painting a Chrome Ball},
  author  = {Phongthawee, Pakkapon and Chinchuthakun, Worameth and Sinsunthithet, Nontaphat and Raj, Aditya and Jampani, Varun and Khungurn, Pramook and Suwajanakorn, Supasorn},
  journal = {arXiv preprint arXiv:2303.13009},
  year    = {2023}
}


%% file: spar3d.bib
@article{oquab2023dinov2,
  title   = {Dinov2: Learning robust visual features without supervision},
  author  = {Oquab, Maxime and Darcet, Timoth{\'e}e and Moutakanni, Th{\'e}o and Vo, Huy and Szafraniec, Marc and Khalidov, Vasil and Fernandez, Pierre and Haziza, Daniel and Massa, Francisco and El-Nouby, Alaaeldin and others},
  journal = {arXiv preprint arXiv:2304.07193},
  year    = {2023}
}

@article{chang2015shapenet,
  title   = {Shapenet: An information-rich 3d model repository},
  author  = {Chang, Angel X and Funkhouser, Thomas and Guibas, Leonidas and Hanrahan, Pat and Huang, Qixing and Li, Zimo and Savarese, Silvio and Savva, Manolis and Song, Shuran and Su, Hao and others},
  journal = {arXiv preprint arXiv:1512.03012},
  year    = {2015}
}

@article{deitke2022objaverse,
  title   = {Objaverse: A Universe of Annotated 3D Objects},
  author  = {Deitke, Matt and Schwenk, Dustin and Salvador, Jordi and Weihs, Luca and Michel, Oscar and VanderBilt, Eli and Schmidt, Ludwig and Ehsani, Kiana and Kembhavi, Aniruddha and Farhadi, Ali},
  journal = {arXiv preprint arXiv:2212.08051},
  year    = {2022}
}

@inproceedings{reizenstein2021co3d,
  title     = {Common objects in 3d: Large-scale learning and evaluation of real-life 3d category reconstruction},
  author    = {Reizenstein, Jeremy and Shapovalov, Roman and Henzler, Philipp and Sbordone, Luca and Labatut, Patrick and Novotny, David},
  booktitle = {Proceedings of the IEEE/CVF International Conference on Computer Vision},
  pages     = {10901--10911},
  year      = {2021}
}

@inproceedings{downs2022google,
  title        = {Google scanned objects: A high-quality dataset of 3d scanned household items},
  author       = {Downs, Laura and Francis, Anthony and Koenig, Nate and Kinman, Brandon and Hickman, Ryan and Reymann, Krista and McHugh, Thomas B and Vanhoucke, Vincent},
  booktitle    = {2022 International Conference on Robotics and Automation (ICRA)},
  pages        = {2553--2560},
  year         = {2022},
  organization = {IEEE}
}

@inproceedings{shue20233d,
  title     = {3d neural field generation using triplane diffusion},
  author    = {Shue, J Ryan and Chan, Eric Ryan and Po, Ryan and Ankner, Zachary and Wu, Jiajun and Wetzstein, Gordon},
  booktitle = {Proceedings of the IEEE/CVF Conference on Computer Vision and Pattern Recognition},
  pages     = {20875--20886},
  year      = {2023}
}

@inproceedings{cheng2023sdfusion,
  title     = {Sdfusion: Multimodal 3d shape completion, reconstruction, and generation},
  author    = {Cheng, Yen-Chi and Lee, Hsin-Ying and Tulyakov, Sergey and Schwing, Alexander G and Gui, Liang-Yan},
  booktitle = {Proceedings of the IEEE/CVF Conference on Computer Vision and Pattern Recognition},
  pages     = {4456--4465},
  year      = {2023}
}

@inproceedings{yariv2024mosaic,
  title     = {Mosaic-sdf for 3d generative models},
  author    = {Yariv, Lior and Puny, Omri and Gafni, Oran and Lipman, Yaron},
  booktitle = {Proceedings of the IEEE/CVF Conference on Computer Vision and Pattern Recognition},
  pages     = {4630--4639},
  year      = {2024}
}

@inproceedings{liu2023zero,
  title     = {Zero-1-to-3: Zero-shot one image to 3d object},
  author    = {Liu, Ruoshi and Wu, Rundi and Van Hoorick, Basile and Tokmakov, Pavel and Zakharov, Sergey and Vondrick, Carl},
  booktitle = {Proceedings of the IEEE/CVF International Conference on Computer Vision},
  pages     = {9298--9309},
  year      = {2023}
}

@article{shi2023mvdream,
  title   = {Mvdream: Multi-view diffusion for 3d generation},
  author  = {Shi, Yichun and Wang, Peng and Ye, Jianglong and Long, Mai and Li, Kejie and Yang, Xiao},
  journal = {arXiv preprint arXiv:2308.16512},
  year    = {2023}
}

@inproceedings{mescheder2019occupancy,
  title     = {Occupancy networks: Learning 3d reconstruction in function space},
  author    = {Mescheder, Lars and Oechsle, Michael and Niemeyer, Michael and Nowozin, Sebastian and Geiger, Andreas},
  booktitle = {Proceedings of the IEEE/CVF Conference on Computer Vision and Pattern Recognition},
  pages     = {4460--4470},
  year      = {2019}
}

@inproceedings{choy20163d,
  title        = {3d-r2n2: A unified approach for single and multi-view 3d object reconstruction},
  author       = {Choy, Christopher B and Xu, Danfei and Gwak, JunYoung and Chen, Kevin and Savarese, Silvio},
  booktitle    = {European conference on computer vision},
  pages        = {628--644},
  year         = {2016},
  organization = {Springer}
}

@inproceedings{wang2018pixel2mesh,
  title     = {Pixel2mesh: Generating 3d mesh models from single rgb images},
  author    = {Wang, Nanyang and Zhang, Yinda and Li, Zhuwen and Fu, Yanwei and Liu, Wei and Jiang, Yu-Gang},
  booktitle = {Proceedings of the European conference on computer vision (ECCV)},
  pages     = {52--67},
  year      = {2018}
}

@inproceedings{liu2019soft,
  title     = {Soft rasterizer: A differentiable renderer for image-based 3d reasoning},
  author    = {Liu, Shichen and Li, Tianye and Chen, Weikai and Li, Hao},
  booktitle = {Proceedings of the IEEE/CVF International Conference on Computer Vision},
  pages     = {7708--7717},
  year      = {2019}
}

@article{hong2023lrm,
  title   = {Lrm: Large reconstruction model for single image to 3d},
  author  = {Hong, Yicong and Zhang, Kai and Gu, Jiuxiang and Bi, Sai and Zhou, Yang and Liu, Difan and Liu, Feng and Sunkavalli, Kalyan and Bui, Trung and Tan, Hao},
  journal = {arXiv preprint arXiv:2311.04400},
  year    = {2023}
}

@article{TripoSR2024,
  title   = {TripoSR: Fast 3D Object Reconstruction from a Single Image},
  author  = {Tochilkin, Dmitry and Pankratz, David and Liu, Zexiang and Huang, Zixuan and and Letts, Adam and Li, Yangguang and Liang, Ding and Laforte, Christian and Jampani, Varun and Cao, Yan-Pei},
  journal = {arXiv preprint arXiv:2403.02151},
  year    = {2024}
}

@article{tang2024lgm,
  title   = {Lgm: Large multi-view gaussian model for high-resolution 3d content creation},
  author  = {Tang, Jiaxiang and Chen, Zhaoxi and Chen, Xiaokang and Wang, Tengfei and Zeng, Gang and Liu, Ziwei},
  journal = {arXiv preprint arXiv:2402.05054},
  year    = {2024}
}

@article{sf3d2024,
  title   = {SF3D: Stable Fast 3D Mesh Reconstruction with UV-unwrapping and Illumination Disentanglement},
  author  = {Boss, Mark and Huang, Zixuan and Vasishta, Aaryaman and Jampani, Varun},
  journal = {arXiv preprint},
  year    = {2024}
}

@article{wang2024prolificdreamer,
  title   = {Prolificdreamer: High-fidelity and diverse text-to-3d generation with variational score distillation},
  author  = {Wang, Zhengyi and Lu, Cheng and Wang, Yikai and Bao, Fan and Li, Chongxuan and Su, Hang and Zhu, Jun},
  journal = {Advances in Neural Information Processing Systems},
  volume  = {36},
  year    = {2024}
}

@article{gardner2023reni++,
  title   = {RENI++ A Rotation-Equivariant, Scale-Invariant, Natural Illumination Prior},
  author  = {Gardner, James AD and Egger, Bernhard and Smith, William AP},
  journal = {arXiv preprint arXiv:2311.09361},
  year    = {2023}
}


%% file: spv.bib
@inproceedings{poole2022dreamfusion,
  title     = {{DreamFusion}: Text-to-{3D} using {2D} Diffusion},
  author    = {Poole, Ben and Jain, Ajay and Barron, Jonathan T and Mildenhall, Ben},
  booktitle = {ICLR},
  year      = {2023}
}

@article{kerbl20233d,
  title   = {{3D} gaussian splatting for real-time radiance field rendering},
  author  = {Kerbl, Bernhard and Kopanas, Georgios and Leimk{\"u}hler, Thomas and Drettakis, George},
  journal = {ACM Transactions on Graphics},
  year    = {2023}
}

@article{mildenhall2021nerf,
  title   = {{NeRF}: Representing scenes as neural radiance fields for view synthesis},
  author  = {Mildenhall, Ben and Srinivasan, Pratul P and Tancik, Matthew and Barron, Jonathan T and Ramamoorthi, Ravi and Ng, Ren},
  journal = {Communications of the ACM},
  year    = {2021}
}

@inproceedings{voleti2024sv3d,
  author    = {Voleti, Vikram and Yao, Chun-Han and Boss, Mark and Letts, Adam and Pankratz, David and Tochilkin,  Dmitrii and Laforte, Christian and Rombach, Robin and Jampani, Varun},
  title     = {{SV3D}: Novel Multi-view Synthesis and {3D} Generation from a Single Image using Latent Video Diffusion},
  booktitle = {ECCV},
  year      = {2024}
}

@inproceedings{long2023wonder3d,
  title     = {Wonder{3D}: Single image to {3D} using cross-domain diffusion},
  author    = {Long, Xiaoxiao and Guo, Yuan-Chen and Lin, Cheng and Liu, Yuan and Dou, Zhiyang and Liu, Lingjie and Ma, Yuexin and Zhang, Song-Hai and Habermann, Marc and Theobalt, Christian and others},
  booktitle = {CVPR},
  year      = {2024}
}

@article{tochilkin2024triposr,
  title   = {{TripoSR}: Fast {3D} object reconstruction from a single image},
  author  = {Tochilkin, Dmitry and Pankratz, David and Liu, Zexiang and Huang, Zixuan and Letts, Adam and Li, Yangguang and Liang, Ding and Laforte, Christian and Jampani, Varun and Cao, Yan-Pei},
  journal = {arXiv preprint arXiv:2403.02151},
  year    = {2024}
}

@article{xiang2024trellis,
  title   = {Structured 3D Latents for Scalable and Versatile 3D Generation},
  author  = {Xiang, Jianfeng and Lv, Zelong and Xu, Sicheng and Deng, Yu and Wang, Ruicheng and Zhang, Bowen and Chen, Dong and Tong, Xin and Yang, Jiaolong},
  journal = {arXiv preprint arXiv:2412.01506},
  year    = {2024}
}

@inproceedings{vainer2024collaborative,
  title     = {Collaborative control for geometry-conditioned {PBR} image generation},
  author    = {Vainer, Shimon and Boss, Mark and Parger, Mathias and Kutsy, Konstantin and De Nigris, Dante and Rowles, Ciara and Perony, Nicolas and Donn{\'e}, Simon},
  booktitle = {ECCV},
  year      = {2024}
}

@article{zhao2025hunyuan3d,
  title   = {Hunyuan3d 2.0: Scaling diffusion models for high resolution textured 3d assets generation},
  author  = {Zhao, Zibo and Lai, Zeqiang and Lin, Qingxiang and Zhao, Yunfei and Liu, Haolin and Yang, Shuhui and Feng, Yifei and Yang, Mingxin and Zhang, Sheng and Yang, Xianghui and others},
  journal = {arXiv preprint arXiv:2501.12202},
  year    = {2025}
}
